\documentclass[10pt,journal,compsoc,twocolumn]{IEEEtran}

\usepackage{amsmath,amssymb}
\usepackage{graphicx}
\usepackage{booktabs}
\usepackage{multirow}
\usepackage{array}
\usepackage{caption}
\usepackage{subcaption}
\usepackage{xcolor}
\usepackage{hyperref}
\usepackage{etoolbox}
\usepackage{url}
\usepackage[numbers,sort&compress]{natbib}

\AtBeginEnvironment{table}{\small}
\newcommand{\fitcol}[1]{%
  \resizebox{\ifdim\width>\columnwidth\columnwidth\else\width\fi}{!}{#1}}

\definecolor{accentblue}{RGB}{31,119,180}
\definecolor{accentorange}{RGB}{255,127,14}
\definecolor{accentgreen}{RGB}{44,160,44}

\title{Label-Space Mismatch Distorts Cross-Cohort Evaluation in Breast Cancer Molecular Subtyping}



\author{Suxing Liu\textsuperscript{1,*}, Byungwon Min\textsuperscript{1}, Zhijian Zheng\textsuperscript{2}, and Jiahao Wang\textsuperscript{3}
\thanks{*Correspondence: bentondoucet@gmail.com}%
\thanks{S. Liu and B. Min are with the Department of IT Engineering, Mokwon University, Daejeon 35349, Republic of Korea}%
\thanks{S. Liu is also with the School of Digital Media Technology, Jiangxi Arts \& Ceramics Technology Institute, Jingdezhen 333001, China}%
\thanks{Z. Zheng is with the University of Alabama, Tuscaloosa, AL 35487, USA}%
\thanks{J. Wang is with the School of Computer Sciences, Universiti Sains Malaysia, Minden Heights, Penang 11800, Malaysia}%
}

\begin{document}

\maketitle

\begin{abstract}
Cross-cohort evaluation of breast cancer subtype classifiers is unreliable when source and target cohorts use different annotation standards: METABRIC labels six classes, PAM50 defines five, and a claudin-low prediction cannot be correct under a five-class standard. A classifier transferred from 1{,}369 METABRIC patients (train n=958) to TCGA-BRCA (n=987) and GSE21653 (n=266) separates three evaluation operations routinely conflated in cross-site comparisons. \textbf{(1) Label-space restriction} -- an evaluation-rule change, not a retraining gain -- is the dominant correction (73.76\%$\rightarrow$79.74\% on TCGA-BRCA; 63.16\%$\rightarrow$68.42\% on GSE21653); \textbf{(2) EM prior correction} on the reconciled space is an optional calibration step (Brier fidelity up, macro F1 down; to 80.65\%% and 74.06\%%); and \textbf{(3)} the target label space is \emph{not recoverable} from unlabelled predictions (EM retains prior 0.0751 on the absent class; calibrated detectors fail), so compliance must come from cohort documentation.

The protocol exposes why internal validation overstates cross-cohort value: multimodal fusion gains do not transfer (no configuration beats mRNA alone, n=694; McNemar $p$=0.13--1.00), and the genomic panel adds nothing to pathological complete response prediction over clinical variables ($\Delta$AUC = +0.0045). Clinically, without label-space compliance, cross-site accuracy comparisons misattribute annotation-standard differences to model skill; Luminal A/B and claudin-low boundary errors map to distinct treatment decisions; and subtype accuracy is a weaker portable endpoint than survival discrimination (external C-index 0.577--0.584). Reporting the cohort's allowed label set and applying label-space compliance are therefore prerequisites before any cross-cohort performance claim.
\end{abstract}


\section{Introduction}
Breast cancer is a heterogeneous disease characterized by distinct molecular subtypes that differ in gene expression patterns, clinical behavior, and treatment response \cite{perou2000molecular}. The PAM50 classification system defines five intrinsic subtypes: Luminal A, Luminal B, HER2-enriched, Basal-like, and Normal-like \cite{parker2009supervised}. Claudin-low is a separate designation based on additional markers \cite{prat2010phenotypic}. Accurate subtyping can inform treatment selection and prognosis assessment \cite{prat2014research}.

Current approaches to breast cancer subtyping predominantly rely on single data modalities. The Prosigna (PAM50) test uses only gene expression data \cite{parker2009supervised}, while clinical risk calculators such as Nottingham Prognostic Index use only clinicopathological features \cite{haybittle1982prognostic}. However, breast cancer biology is driven by the interplay of genomic, epigenomic, and clinical factors. DNA methylation patterns, for instance, play a crucial role in regulating gene expression in breast cancer \cite{pasculli2018epigenetics}, and clinical variables such as ER/PR/HER2 status carry independent prognostic information \cite{goldhirsch2013personalizing}.

Multimodal learning has emerged as a promising paradigm for cancer classification, with studies demonstrating that integrating multiple data types can improve prediction accuracy and biological understanding \cite{wang2014similarity}. However, several challenges remain: (1) many multimodal approaches require large training datasets that are unavailable for rare cancer subtypes; (2) models trained on one profiling platform often fail to generalize to data from different platforms; and (3) the relative contributions of different modalities are poorly understood.

Label-shift methods provide established tools for estimating or correcting changes in class prevalence from unlabeled target predictions. These include black-box shift estimation, EM-based prior adjustment, regularized adaptation, calibration-aware estimation, and methods for open-set or generalized label shift \cite{saerens2002adjusting}. These methods address changes in class proportions within a specified label space; they do not by themselves establish that source and target annotation standards define the same set of classes. That gap is a silent confounder in every cross-cohort subtyping comparison, and we make closing it a methodologically explicit operation -- the evaluation protocol this paper formalizes -- applied to breast cancer molecular-subtype classifiers rather than proposing yet another fusion or label-shift algorithm.

We use a multimodal METABRIC model as an internal reference and focus the cross-cohort analysis on its mRNA classifier, the feature block available across cohorts. The central question is whether apparent performance differences can be interpreted before reconciling the target cohort's annotation space. We quantify the effect of restricting predictions to the five PAM50 classes, separate that operation from prior correction, and test whether absent classes can be inferred from unlabeled predictions. Survival discrimination and multimodal feature attribution provide secondary analyses.

The analysis raised a question about a lower accuracy observed in one external cohort. The target cohort used PAM50 annotations, whereas the METABRIC model had six output classes, including claudin-low. Predictions assigned to a class excluded by the target annotation standard cannot be correct under that evaluation -- and the errors this produces are clinically consequential: a Luminal A/B boundary error can shift a chemo-endocrine decision, and an un-reconciled claudin-low assignment mislabels a basal-like tumour as a separate class the target standard does not recognise. Section~\ref{sec:labelspace} measures the effect of reconciling the label space and compares it with prior correction. It also evaluates whether the absent class can be identified from unlabeled target predictions.

Our key contributions are:
\begin{enumerate}
  \item \textbf{A general protocol for label-space compliance in cross-cohort comparison.} We formalize the evaluation operation that cross-cohort subtyping models routinely omit -- restricting a source-trained classifier's output to the target's allowed label set, separating that restriction from class-prior correction, and testing whether the absent label can be inferred from unlabeled target predictions -- as a protocol with explicit inputs (source training space, target annotation standard, trained classifier) and outputs (a compliant cross-cohort accuracy and its allocation across label-space and prior effects). We instantiate it on breast cancer molecular subtypes: transferring a six-class METABRIC classifier to five-class PAM50 cohorts, reconciling the label space and correcting priors moves TCGA-BRCA from 73.76\% to 80.65\% and GSE21653 from 63.16\% to 74.06\% without retraining (Section~\ref{sec:labelspace}). The protocol is not specific to any one absent class: the controlled simulation in Section~\ref{sec:absence_generality} applies it to every absent class across three absence modes and shows the correction tracks the absent class's prior mass rather than a particular annotation;
  \item \textbf{Empirical limits of label-free recovery.} Unconstrained EM retains an estimated prior of 0.0751 on an absent class (claudin-low), and neither a calibrated geometric detector nor a generative model-selection procedure recovers the target annotation space under covariate and prior shift. The practical consequence is that target label sets must be specified from cohort documentation, not inferred from predictions;
  \item \textbf{Protocol-validated evidence on where cross-cohort value does and does not transfer.} Applying the protocol in a harmonized space shared by METABRIC and TCGA-BRCA (n=694), no multimodal configuration beat mRNA alone (McNemar $p$=0.13--1.00), and the best internal model rested on 59 METABRIC-specific clinical dimensions with no external counterpart. The protocol converts this from a report of failure into a deployment rule: modality and architecture choices validated on a single cohort are weak evidence of cross-cohort value, the portable endpoint is survival discrimination (C-index 0.577--0.584) rather than subtype accuracy, and annotation-standard misreads -- Luminal A/B and claudin-low boundary errors -- map to distinct treatment decisions;
  \item \textbf{Interpretable, clinically relevant secondary analyses.} Exact SHAP attribution (mRNA 63.3\%, methylation 26.7\%, clinical 10.0\%), significant prognostic stratification (C-index 0.598, $p=0.0008$), and a pCR analysis (GSE20194, n=278) showing the genomic panel adds nothing over clinical variables ($\Delta$AUC = +0.0045) complete an evaluation that quantifies \emph{where} and \emph{how} multimodal subtyping decisions can be trusted across cohorts.
\end{enumerate}

\section{Related Work}
\label{sec:related}

\subsection{Multimodal and Multi-Omics Learning in Cancer}
Fusing heterogeneous molecular and clinical data is now standard in computational oncology, with deep architectures applied to diagnosis, prognosis and treatment selection \cite{moon2023machine,jiang2024autosurv}. The dominant recent design is graph-based, representing multi-omics profiles as graphs combined through attention or message passing \cite{lin2026moagnn,fan2026cmgl,tan2025amogel,dou2025mogcan,zhang2025hallmarkgraph,tanvir2024mogat,xiao2023graph,zhang2025multiomics}, often with hierarchical or multi-view objectives \cite{yang2024multimodal}; alternatives include tensor and stacked fusion \cite{zhang2024msfn}, contrastive fusion \cite{liu2023attentionbased}, and autoencoder joint embedding \cite{benkirane2023customics}. In breast cancer, attention-based fusion has been applied to neoadjuvant chemotherapy response \cite{zhang2025attentionbased}, and multimodal systems to pathological complete response and survival \cite{mao2025a,cheng2026aligned,jia2026dynamic}. Such methods are typically assessed by internal cross-validation, and recent reviews note that fusion architectures are rarely validated on a cohort drawn from a different source, and that the field lacks shared evaluation protocols \cite{steyaert2023multimodal,kibria2026decoding}. Our study is complementary: we propose no new fusion architecture, and the one we tested did not beat a linear reference (Section~\ref{sec:discussion}). We examine instead what happens to the \emph{evaluation} of such models when the target cohort's annotation standard differs from the source's.

\subsection{Label Shift, Domain Adaptation and Calibration}
Of the distribution-shift literature, the closest strand is label shift, in which class proportions change while class-conditional distributions do not. Early work estimated target priors from unlabelled predictions by expectation-maximisation \cite{saerens2002adjusting}, of which our procedure is a direct descendant; subsequent work refined estimation and correction \cite{stan2024unsupervised}, extended it to open-set and generalised settings where the label spaces themselves may differ \cite{yang2025a1,yoshida2025classprior,volpi2025leap}, and connected it to test-time and source-free adaptation \cite{chen2025vpsfda,lai2023domainaware}. This literature treats the label space as given and estimates only priors within it; open-set label shift is the nearest exception \cite{yang2025a1} but still assumes the target's label space is discoverable from data. Our negative result speaks to that assumption: when a class is genuinely unrepresentable under the target's annotation standard, the tested prior-estimation and detection procedures did not recover the fact, so the correction must be supplied from cohort documentation rather than inferred. Calibration is the companion question -- whether corrected probabilities remain faithful to observed frequencies -- which we address with Brier and ECE in the Results.

\subsection{PAM50 Annotation and Cross-Platform Concordance}
The intrinsic subtypes are defined by a 50-gene signature \cite{parker2009supervised} whose assignment is not perfectly reproducible: studies report discordance with immunohistochemistry-based surrogates \cite{sayed2026immunohistochemistry}, between primary and metastatic lesions \cite{jrgensen2021pam}, and across populations \cite{hohmann2025molecular}, with assay reproducibility limited by intratumoral heterogeneity \cite{hurson2023reproducibility}. Methodological work seeks to stabilise calls through rank-based normalisation \cite{kandan2025rank}, improved implementations \cite{rajkumar2025enhanced}, and transferable derived scores \cite{denegreirosbotan2026corepam}, while batch and preprocessing effects remain an active concern \cite{rojasvelazquez2024methodology,van2024a,ng2023the}. Such work establishes that annotation disagreement is real and consequential, but addresses disagreement within a shared label space -- two assays disagreeing about which of five subtypes a tumour is. The failure we examine is one level up: these cohorts do not share a label space at all, because one annotates a sixth class the other's standard cannot express. Our supplementary sweep shows that this mismatch is governed by the prior mass of the absent class, and that its most common real form -- a class folded into a neighbouring label rather than dropped -- is also its most damaging.

\section{Methods}

\subsection{METABRIC Dataset}
The Molecular Taxonomy of Breast Cancer International Consortium (METABRIC) dataset \cite{curtis2012genomic} was used as the primary development cohort. The source cohort contained 1,974 patients with PAM50 annotations. The analysis set comprised 1,369 patients retained after aligning patient identifiers across the clinical, mRNA, copy-number, methylation, and mutation records used in preprocessing. The 1,369-patient analysis set was split into training, validation, and test subsets. Five-fold cross-validation estimates were computed within the training subset (n=958), not across the full aligned analysis set.

Clinical features were encoded from 22 categorical and continuous variables including age at diagnosis, tumor size, tumor stage, ER/PR/HER2 status, and treatment history. Categorical variables were one-hot encoded, and continuous variables were Z-score standardized, yielding 69 clinical feature dimensions. mRNA expression data were filtered to 412 genes (subset of the 488 available genes matching the TCGA gene panel), and DNA methylation comprised 232 CpG probes selected by variance ranking.

The analysis set was split into training (n=958, 70\%), validation (n=205, 15\%), and test (n=206, 15\%) sets using stratified random sampling to preserve subtype proportions. Class distribution in the training set: Luminal A (35.7\%), Luminal B (23.9\%), HER2-enriched (12.5\%), claudin-low (9.5\%), Basal-like (10.4\%), and Normal-like (7.9\%).

\subsection{TCGA-BRCA External Validation Dataset}
The Cancer Genome Atlas Breast Cancer (TCGA-BRCA) cohort \cite{cancer2012comprehensive} was used for external validation. From the Firehose Legacy study (1,108 patients), we obtained RNA-seq V2 expression data (1,100 samples) and PAM50 annotations (987 samples with 5 subtypes; claudin-low not annotated). Survival data were available for 966 patients with 133 death events (13.8\% event rate).

Gene expression was mapped from the 412-gene METABRIC panel using Entrez Gene IDs via the cBioPortal API. Within each platform, expression values were rank-transformed and normalized using the inverse normal CDF to achieve cross-platform comparability:
\begin{equation}
  x_{ij}^{\text{norm}} = \Phi^{-1}\left(\frac{\text{rank}(x_{ij})}{n_j + 1}\right)
  \label{eq:rank_norm}
\end{equation}
where $\Phi^{-1}$ is the inverse normal CDF and $n_j$ is the number of samples in platform $j$. This within-platform normalization preserves relative sample ordering while removing platform-specific distributional differences.

\subsection{GEO External Cohorts}
GSE21653 was obtained from the Gene Expression Omnibus (GEO). It comprises 266 tumors profiled on GPL570 (Affymetrix Human Genome U133 Plus 2.0), with PAM50 subtype calls and disease-free survival records. Of the 412 genes in the METABRIC panel, 369 mapped to this platform. When multiple probes mapped to one gene, the probe with the highest mean expression was retained. Expression values were normalized within the cohort using the rank-based procedure above.

GSE20194 was obtained from GEO for the secondary pathological complete response (pCR) analysis. The cohort included 278 patients treated with neoadjuvant chemotherapy, of whom 56 achieved pCR and 222 had residual disease. The series was profiled on the Affymetrix Human Genome U133A platform (GPL96); 276 genes from the 412-gene panel mapped to this platform. The published pCR labels were used only as the outcome for this secondary analysis.

\subsection{Multimodal Feature Engineering}
We engineered features from five modalities:
\begin{enumerate}
  \item \textbf{Clinical} (69d): One-hot encoded categorical variables (e.g., ER status, HER2 status, tumor stage, surgery type, treatment history) with Z-score standardized continuous variables (age, tumor size, lymph node count, NPI).
  \item \textbf{mRNA expression} (412d): Microarray expression values for 412 genes, Z-score standardized.
  \item \textbf{Copy number alteration} (256d): Top 256 genes by variance from SNP 6.0 array, Z-score standardized.
  \item \textbf{DNA methylation} (232d): Beta values for 232 CpG probes selected by variance, Z-score standardized.
  \item \textbf{Somatic mutations} (173d): Binary indicators for 173 genes with at least one mutation in the cohort.
\end{enumerate}

Total feature dimensions across all modalities: $69 + 412 + 256 + 232 + 173 = 1{,}142$.

\subsection{Model Development}
Multinomial logistic regression was chosen as the primary classifier due to its strong performance on small-sample, high-dimensional data and inherent interpretability. The model uses the softmax function:
\begin{equation}
  P(y = k \mid \mathbf{x}) = \frac{\exp(\mathbf{w}_k^\top \mathbf{x} + b_k)}{\sum_{j=1}^{K} \exp(\mathbf{w}_j^\top \mathbf{x} + b_j)}
  \label{eq:softmax}
\end{equation}
with $L_2$ regularization controlled by parameter $C$. We evaluated $C \in \{0.001, 0.01, 0.1, 1.0, 10.0\}$, with $C = 0.01$ selected by cross-validation. Estimators were implemented with scikit-learn \cite{pedregosa2011scikit}.

\subsection{Comparison, Ensemble and Transformer Variants}
We compared logistic-regression regularization variants and a sigmoid-calibrated model with Random Forest (300 trees), XGBoost, LightGBM, CatBoost, Extra Trees, and soft-vote ensembles \cite{vaswani2017attention}. Ensemble methods included a soft-vote of LogReg+GBDT+XGBoost and a two-stage cascade sending low-confidence samples to a complementary model. A cross-modal fusion Transformer used five modality-specific linear projections into a two-layer encoder ($d_{\text{model}}=64$, four attention heads) with a 174K-parameter classification head; we report its negative result in the Discussion.

\subsection{Evaluation Protocol}
Primary metrics: accuracy and macro-averaged F1-score; secondary metrics are per-class precision, recall, and F1. Statistical significance of model differences was assessed using McNemar's test on the held-out test set.

\subsection{Survival Analysis}
Predicted subtype probabilities were used as covariates in a Cox proportional hazards model:
\begin{equation}
  h(t) = h_0(t) \exp\left(\sum_{k=1}^{K-1} \beta_k \hat{p}_k\right)
  \label{eq:cox}
\end{equation}
where $\hat{p}_k$ is the predicted probability for subtype $k$ (Luminal A as reference), and $h_0(t)$ is the baseline hazard. Discrimination was assessed using the concordance index (C-Index) at 3-year, 5-year, and 10-year time horizons. Kaplan-Meier survival curves were compared using the log-rank test.

\subsection{Interpretability Analysis}
SHAP (SHapley Additive exPlanations) \cite{lundberg2017unified} with the LinearExplainer was used to quantify feature importance. The explainer is exact rather than approximate here: the fitted classifier is multinomial logistic regression, a linear model, and for a linear model the LinearExplainer computes the Shapley values in closed form from the coefficients and the background distribution. This is not a convenience fallback for a nonlinear model but the estimator matched to the hypothesis class. Modality contribution was computed as the mean absolute SHAP value aggregated across all features within each modality. The analysis was performed on a 300-sample subset of the training data.

\subsection{Label-Space Reconciliation and Prior Correction}
\label{sec:methods_shift}

Because the training cohort carries six classes and the external cohorts carry five, the classifier's output space must be reconciled before any accuracy is computed. We define a \emph{label-space mask} $m \in \{0,1\}^K$ with $m_k = 0$ for any class $k$ that the target cohort's annotation standard cannot emit. For PAM50-annotated cohorts this mask is fixed by the annotation standard alone. Claudin-low is not one of the five PAM50 intrinsic subtypes \cite{parker2009supervised}, so the mask is \emph{independent of the target labels}. This distinction is load-bearing and we state it explicitly: deriving the mask from $\text{unique}(y_{\text{target}})$ would leak the labels and inflate the result, so every number reported in Section~\ref{sec:labelspace} uses the hard-coded annotation mask.

We compare four estimators within each target cohort. They share the same fitted classifier and feature set within that cohort. TCGA-BRCA uses all 412 mapped genes. GSE21653 uses the 369 genes mapped to GPL570, with the source classifier refitted on the corresponding METABRIC columns:

\begin{enumerate}
  \item \textbf{base}: $\arg\max_k p_k$, with no reconciliation (the conventional protocol);
  \item \textbf{mask\_only}: zero the masked columns and renormalise;
  \item \textbf{em\_pam50}: expectation-maximisation prior correction \cite{saerens2002adjusting} over the full $K$-class space, with masked columns fixed at zero throughout;
  \item \textbf{em\_full}: the same EM procedure with \emph{no} mask, meaning it receives no label-space knowledge whatsoever.
\end{enumerate}

The source prior is the METABRIC training composition. EM updates the target probabilities and class priors as follows:
\begin{align*}
p_k^{(t+1)}(x) &\propto p_k^{(t)}(x)\, \pi_k^{(t)} / \pi_k^{(0)}, \\
\pi^{(t+1)} &= \mathbb{E}_x[p^{(t+1)}(x)].
\end{align*}
Iterations stop when the prior changes by less than $10^{-6}$. The contrast between \textbf{em\_pam50} and \textbf{em\_full} isolates the value of the label-space fact. Both use identical unlabelled target predictions, but only one is given the structurally impossible classes.

Gains are reported with a paired bootstrap ($n_{\text{iter}} = 2000$, resampling target samples and recomputing both endpoints), and a split-half transductive check confirms the correction does not depend on how the target is partitioned.

\subsection{Controlled Simulation of Mismatch}
\label{sec:methods_sim}

To separate the contributions of label-space mismatch from ordinary prior shift we simulate targets under a full factorial design: label-space mismatch $n_{\text{absent}} \in \{0,1,2\}$, covariate shift $\in \{0.0, 0.4, 0.8\}$, and prior shift $\in \{\text{false}, \text{true}\}$, with 20 seeds per cell and $n_{\text{target}} = 1000$. A class is treated as absent when its target prevalence falls below $\tau_{\text{absent}} = 0.02$ or below $\tau_{\text{ratio}} = 0.10$ of its source prevalence. Each cell reports accuracy with no correction, with an oracle label-space mask, and with a mask estimated from the target.

That design removes only the rarest classes and only removes them completely, which is the pattern realised by claudin-low. Because a reader should not have to assume that the mildest instance is representative, we run a second, larger sweep over two further axes (described in the Results below). The first varies \emph{which} class is absent across all six and all fifteen pairs, so a mismatch involving a common or confusable subtype is covered. The second varies \emph{how} it is absent, distinguishing three patterns seen in practice: a class dropped outright, a class under-ascertained rather than absent (prevalence scaled to $\tau_{\text{ratio}}$ of source), and a class \emph{merged} into a neighbour's label -- claudin-low recorded as Basal-like, or Luminal B recorded as Luminal A, which is the most common form of annotation disagreement between cohorts.

\subsection{Geometric Absence Detection}
\label{sec:methods_detector}

We also asked whether the absent class could be identified from unlabelled predictions alone, without being told. We built a detector on the geometry of the target probability simplex: for candidate absent set $S$, measure the ratio of observed mass to source-expected mass on the columns in $S$, and flag $S$ when that ratio exceeds a threshold $\tau$. The threshold was calibrated on null data (no truly absent class) at level $\alpha = 0.05$, taking the 95th percentile of the maximum ratio statistic under the null. This calibration is what makes the detector a fair test rather than a tuned one: $\tau^*$ is chosen so that a cohort with no absent class is flagged at most 5\% of the time.

\subsection{Generative Model-Selection Detection}
\label{sec:methods_lrt}

Because a hand-calibrated threshold is the obvious objection to the detector above, we also tested a mechanism that sets no threshold at all. Here the target sample is modelled as $p(x) = \sum_{k \in M} \pi_k \, \mathcal{N}(x ; \mu_k, \Sigma)$, with the component means and the shared covariance estimated from the \emph{source} labels and the mixture weights $\pi$ fitted on the unlabelled \emph{target} by expectation-maximisation over an allowed label space $M$. Candidate label spaces are all subsets of size 4, 5, or 6. This includes every configuration with zero, one, or two absent classes. The winner is the subset minimising the Bayesian information criterion $\text{BIC} = -2\log L + (|M| - 1)\log n$, so the complexity penalty, rather than a tuning constant, is what controls the trade-off; predictions are formed on a 40-component PCA projection of the standardised features. No target labels are used at any point.

\section{Results}

\subsection{Modality Selection Reveals Optimal Feature Combination}
\label{sec:modality}
\label{sec:model_comparison}
\label{sec:feature_selection}
\label{sec:modality}

Systematic evaluation of modality combinations (Table~\ref{tab:modality_ablation}) revealed that adding more modalities did not always improve performance. The clinical + mRNA + methylation model (713 dimensions) achieved 79.0\% validation accuracy, 3.4 percentage points above the full five-modality model (75.6\%). In cross-validation, adding CNA alone or mutation alone to the selected three-modality model reduced accuracy by 2.3 and 2.0 points, respectively. The full five-modality model scored 72.4\%, 2.9 points below the selected model.

\begin{table}[htbp]
\centering
\caption{Modality-combination ablation on the METABRIC validation set (n=205). Five-fold cross-validation results use the training subset (n=958). The clinical + mRNA + methylation model exceeds the all-modality model by 3.4 percentage points on the validation split.}
\label{tab:modality_ablation}
\fitcol{%
\begin{tabular}{lccc}
\toprule
\textbf{Modality Combination} & \textbf{Dim} & \textbf{Val Acc (\%)} & \textbf{5-fold CV (\%)} \\
\midrule
Clinical only & 69 & 63.9 & 64.1 $\pm$ 2.7 \\
mRNA only & 412 & 74.6 & 73.8 $\pm$ 1.6 \\
Methylation only & 232 & 56.6 & 54.7 $\pm$ 2.4 \\
\midrule
Clinical + mRNA & 481 & 75.6 & 74.5 $\pm$ 1.9 \\
Clinical + Methylation & 301 & 68.3 & 65.7 $\pm$ 1.9 \\
mRNA + Methylation & 644 & 74.6 & 73.1 $\pm$ 2.6 \\
\midrule
\textbf{Clinical + mRNA + Methylation} & \textbf{713} & \textbf{79.0} & \textbf{75.3 $\pm$ 2.6} \\
+ CNA (4 modalities) & 969 & 75.6 & 73.0 $\pm$ 1.8 \\
+ Mutation (4 modalities) & 886 & 77.6 & 73.3 $\pm$ 2.0 \\
All 5 modalities & 1142 & 75.6 & 72.4 $\pm$ 2.0 \\
\bottomrule
\end{tabular}}
\end{table}

Logistic regression with $C = 0.01$ achieved the highest 5-fold cross-validation accuracy (75.3\% $\pm$ 2.6\%) among the evaluated models (Table~\ref{tab:model_comparison}). Ensemble methods and Transformer architectures did not surpass the linear model in this comparison. The best Transformer variant achieved 68.5\% cross-validated accuracy.

\begin{table}[htbp]
\centering
\caption{Model comparison using five-fold cross-validation on the METABRIC training subset (n=958). Models use the selected clinical + mRNA + methylation feature set (713 dimensions), unless otherwise stated.}
\label{tab:model_comparison}
\fitcol{%
\begin{tabular}{lc}
\toprule
\textbf{Model} & \textbf{CV Accuracy (\%)} \\
\midrule
Logistic Regression ($C=0.01$) & \textbf{75.3} \\
Logistic Regression ($C=0.1$) & 74.1 \\
Logistic Regression ($C=1.0$) & 72.9 \\
Calibrated LogReg (sigmoid) & 74.4 \\
Random Forest (300 trees) & 72.4 \\
XGBoost & 72.0 \\
LightGBM & 71.0 \\
CatBoost & 73.9 \\
Extra Trees & 71.3 \\
Soft-vote Ensemble (LR+GB+XGB) & 72.6 \\
Transformer ($d$=64, 2-layer) & 68.5 \\
Guided Transformer (LR prompt) & 74.0 \\
LR $\rightarrow$ Guided TF Cascade & 76.2 \\
\bottomrule
\end{tabular}}
\end{table}

The cascade approach (LogReg $\rightarrow$ Guided Transformer) achieved a CV accuracy of 76.2\%, compared with 75.3\% for logistic regression alone. This 0.9-point difference was within the standard deviation of both estimates. On TCGA-BRCA, the 412-gene logistic-regression reference reached 73.8\%, whereas the unguided Transformer reached 66.1\% (see Discussion).

ANOVA-based feature selection showed that mRNA-only cross-validated accuracy plateaued at approximately 200 genes (74.5\%). The top discriminating genes included \textit{ACOT4} ($F$=130.8), \textit{A2ML1} ($F$=102.0), and \textit{ABAT} ($F$=98.5). With clinical and methylation features, 200 mRNA genes achieved 75.5\% cross-validated accuracy, close to the full 412-gene panel.

\subsection{Internal Validation Confirms Robust Performance}
\label{sec:internal}
\label{sec:internal}

On the held-out METABRIC test set (n=206), the 3-modality logistic regression achieved 73.8\% accuracy and 0.683 macro F1-score (Table~\ref{tab:internal_per_class}). The model showed strong performance for Luminal A (F1=0.768), Luminal B (F1=0.796), and HER2-enriched (F1=0.755) subtypes, with lower performance for Normal-like (F1=0.417) which had the smallest sample size (n=16).

\begin{table}[htbp]
\centering
\caption{Per-class classification metrics on the METABRIC test set (n=206).}
\label{tab:internal_per_class}
\fitcol{%
\begin{tabular}{lcccc}
\toprule
\textbf{Subtype} & \textbf{Precision} & \textbf{Recall} & \textbf{F1-score} & \textbf{Support} \\
\midrule
Luminal A & 0.700 & 0.851 & 0.768 & 74 \\
Luminal B & 0.812 & 0.780 & 0.796 & 50 \\
HER2-enriched & 0.741 & 0.769 & 0.755 & 26 \\
Claudin-low & 0.733 & 0.579 & 0.647 & 19 \\
Basal-like & 0.778 & 0.667 & 0.718 & 21 \\
Normal-like & 0.625 & 0.312 & 0.417 & 16 \\
\midrule
\textbf{Macro avg} & \textbf{0.732} & \textbf{0.660} & \textbf{0.683} & 206 \\
Weighted avg & 0.738 & 0.738 & 0.730 & 206 \\
\bottomrule
\end{tabular}}
\end{table}

\begin{figure}[htbp]
\centering
\includegraphics[width=0.66\linewidth]{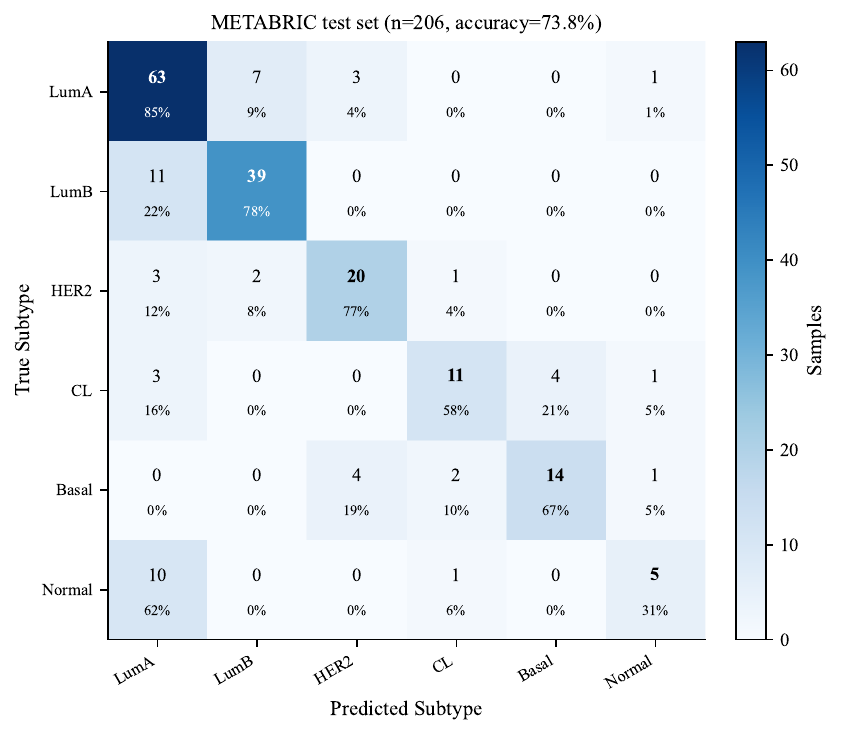}
\caption{Confusion matrix for the six-class PAM50 prediction task on the held-out METABRIC test set (n=206) for the 3-modality logistic regression (Clinical$+$mRNA$+$Methylation, 713 dimensions; accuracy 73.8\%). Cell values are predicted counts with row percentages. Luminal A/Luminal B confusion dominates the error budget.}
\label{fig:confusion}
\end{figure}

Confusion analysis (Figure~\ref{fig:confusion}) revealed that Luminal A/Luminal B misclassification is the primary error source: 9.5\% of Luminal A samples were predicted as Luminal B, and 22.0\% of Luminal B samples were predicted as Luminal A. This reflects the biological similarity between these two subtypes, which share ER-positive status and differ primarily in proliferation rate \cite{prat2014research}. The largest single error rate is on Normal-like, 62.5\% of which is predicted as Luminal A (recall 0.312); with only 16 test samples this estimate is correspondingly imprecise.

\subsection{External Validation Across Distinct Cohorts}
\label{sec:external}
\label{sec:external_multimodal}
\label{sec:external}

\begin{figure}[htbp]
\centering
\includegraphics[width=0.62\linewidth]{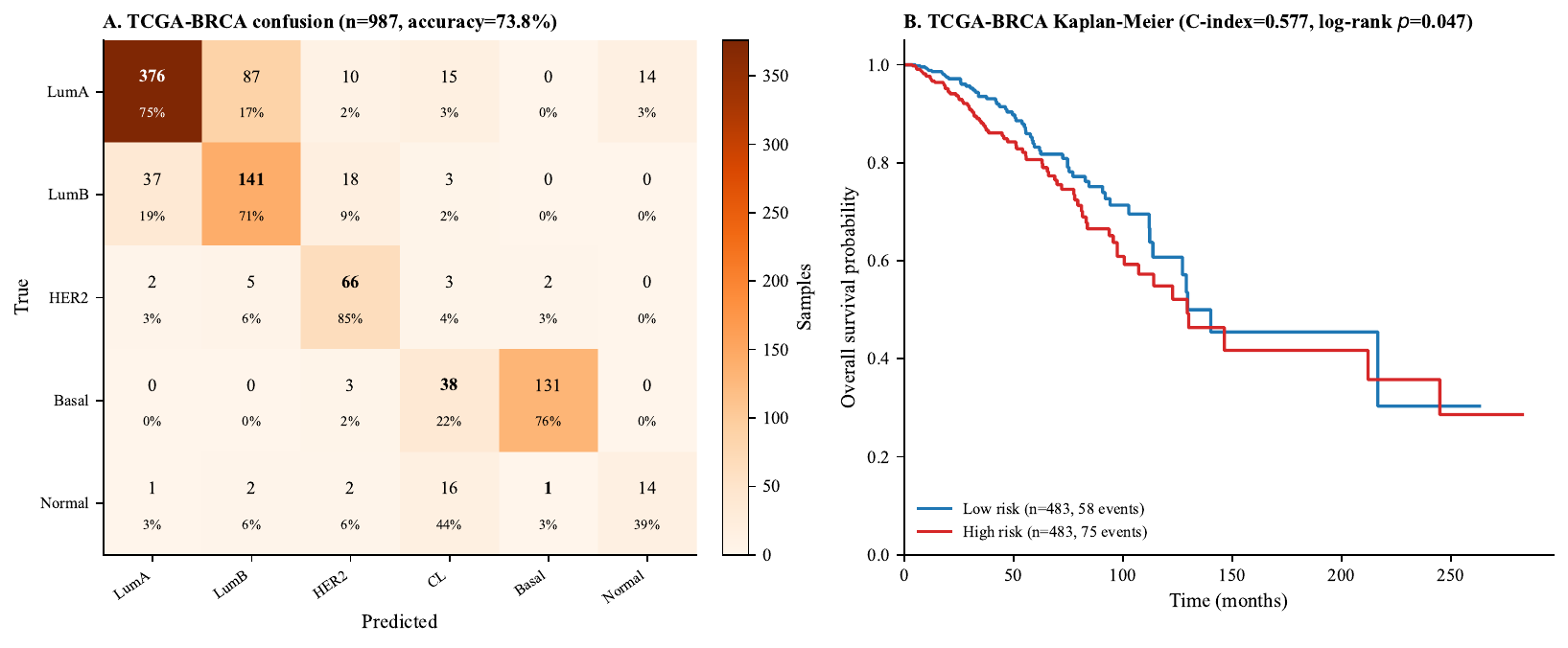}
\caption{External validation on TCGA-BRCA (n=987). The within-platform rank-normalized mRNA model reaches 73.8\% accuracy and a survival C-index of 0.577 ($p$=0.047).}
\label{fig:tcga}
\end{figure}

The 412-gene mRNA-only model with within-platform rank normalization achieved 73.8\% accuracy on TCGA-BRCA (n=987, Figure~\ref{fig:tcga}), similar to its 73.8\% accuracy on the held-out METABRIC test set. The default external evaluation is restricted to mRNA because TCGA's clinical schema is not mappable to METABRIC's in full: it records no Nottingham Prognostic Index or IntClust assignment, has no equivalent of the three-gene classifier or cohort field, and uses different therapy, cellularity and histology vocabularies; only sixteen of the 69 clinical dimensions have a defensible counterpart. A reduced clinical block and a TCGA methylation block can both be constituted, however, and Section~\ref{sec:external_multimodal} uses them to test whether the internal multimodal advantage transfers. The survival C-index was 0.577 ($p = 0.047$, log-rank test).

Macro F1 depends on the set of classes included in the average. We report it over the six-class label space used for training. METABRIC contains all six classes and has macro F1 0.683. TCGA contains five PAM50 classes, so claudin-low contributes an F1 of zero and the six-class average is 0.585. Averaging the same predictions over the five PAM50 classes gives macro F1 \textbf{0.702}. This 0.117 difference is caused by the metric's class set; it does not represent a change in predictions. Separately, the classifier predicted claudin-low for 75 TCGA samples (7.6\%). Those predictions were errors and included 22.1\% of basal-like and 44.4\% of Normal-like samples. TCGA also had a higher Luminal A prevalence than METABRIC (50.9\% versus 36\%) and only 36 Normal-like samples, of which 14 were correctly classified.

The three-modality result in Section~\ref{sec:modality} was established on a single cohort. To test whether it generalizes, we constructed a reduced space that METABRIC and TCGA-BRCA both carry and re-ran the modality ablation inside it. Three layers were harmonized. mRNA uses the same 412-gene panel, matched by symbol on both platforms (all 412 present; 10 are invariant on TCGA and were assigned the neutral score). Methylation uses the 232-gene METABRIC panel, which is already gene-level because the METABRIC preprocessing reduced probes to genes by variance; the corresponding TCGA-BRCA HM450 gene-level values were retrieved from cBioPortal, and 209 of the 232 genes carry measurements (23 are absent from the array). Clinical uses the 10 fields present in both schemas: age, pathologic stage, ER, PR, and HER2 status, and menopausal state. Every continuous feature was rank-transformed to normal scores \emph{within its own cohort}, a label-free operation, so no target labels were used at any point. Requiring all three layers plus an annotated subtype leaves 694 TCGA-BRCA patients across the five PAM50 classes; claudin-low is absent, as in the n=987 cohort.

Table~\ref{tab:multimodal_external} reports the ablation in this space, and the pattern reverses. Adding methylation produces no cross-validated gain on METABRIC (72.91\% for mRNA alone versus 72.91\% for mRNA plus methylation) and lowers TCGA accuracy by 1.9 points, from 80.55\% to 78.67\%. Adding the harmonized clinical block raises METABRIC cross-validated accuracy by 0.5 points and the held-out METABRIC test score by 2.4 points, but moves TCGA accuracy by only 0.3 points in the unrestricted setting and $-0.15$ points under the five-class restriction. No multimodal configuration differs significantly from mRNA alone (paired McNemar, $p = 0.13$ to $1.00$), and the confidence intervals overlap almost completely. These $p$-values are reported without multiplicity adjustment across the five configurations; applying a Bonferroni threshold of $\alpha/5 = 0.01$ leaves every comparison non-significant, so the conclusion is unchanged.

\begin{table}[htbp]
\centering
\caption{Modality ablation in the harmonized three-layer space shared by METABRIC and TCGA-BRCA. CV is five-fold cross-validated accuracy on pooled METABRIC train+val (n=1{,}163). TCGA accuracy is on n=694 patients annotated with one of the five PAM50 classes present in that cohort. The McNemar column is a paired test against mRNA alone. The final row is the original 713-dimension model, shown for reference: its clinical block includes METABRIC-specific fields and has no external counterpart.}
\label{tab:multimodal_external}
\fitcol{%
\begin{tabular}{lcccccc}
\toprule
\textbf{Configuration} & \textbf{Dims} & \textbf{METABRIC CV} & \textbf{METABRIC test} & \textbf{TCGA} & \textbf{TCGA macro F1} & \textbf{$p$} \\
\midrule
mRNA                 & 412 & 72.91 & 72.82 & \textbf{80.55} & 0.768 & --- \\
mRNA + clinical      & 422 & 73.43 & \textbf{75.24} & 80.40 & \textbf{0.774} & 1.00 \\
mRNA + methylation   & 644 & 72.91 & 72.33 & 78.67 & 0.747 & 0.13 \\
clinical + mRNA + methylation & 654 & 73.17 & 74.76 & 79.11 & 0.760 & 0.30 \\
\midrule
\emph{original flagship} & \emph{713} & \emph{75.92} & \emph{73.79} & \emph{n/a} & \emph{n/a} & \emph{---} \\
\bottomrule
\end{tabular}}
\end{table}

\begin{figure}[htbp]
\centering
\includegraphics[width=0.74\linewidth]{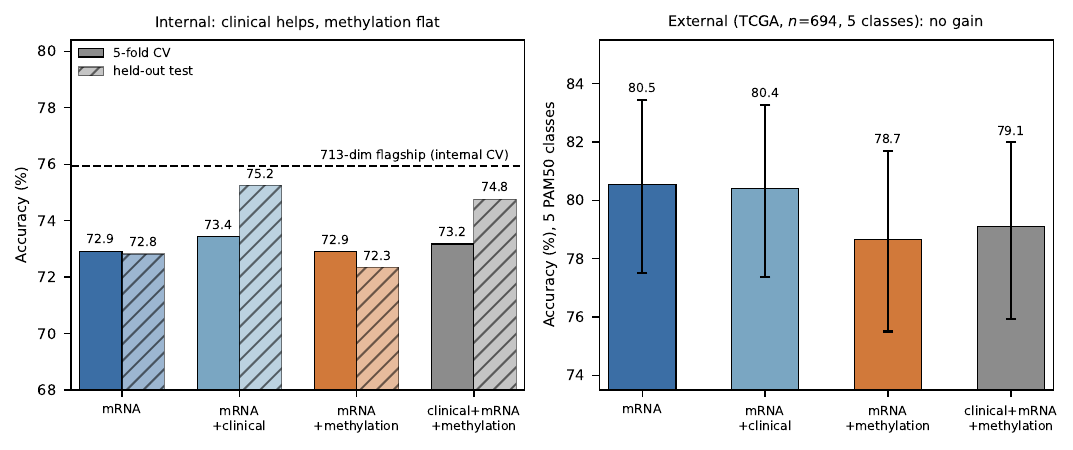}
\caption{The multimodal advantage does not transfer. Panel A: internal cross-validated and held-out METABRIC accuracy for each configuration, with the original 713-dimension flagship as a reference line. Panel B: TCGA-BRCA accuracy over the five PAM50 classes present (n=694) with bootstrap 95\% confidence intervals. Every configuration's interval overlaps the mRNA-only interval; the harmonized clinical block helps internally and is negligible externally.}
\label{fig:multimodal_external}
\end{figure}

The two internal blocks move in opposite directions from their external counterparts. On METABRIC the harmonized tri-modal model reaches 73.17\% cross-validated accuracy, while the original 713-dimension flagship reaches 75.92\%. That 2.75-point gap is attributable to the 59 clinical dimensions that exist only in METABRIC (NPI, GRADE, INTCLUST membership, THREEGENE proliferation class, and cohort indicator); those fields are not available in the target cohort, so the model that performs best internally is precisely the one that cannot be transported. Conversely, the harmonized model's higher absolute TCGA accuracy (80.55\% versus the published 73.76\%) reflects the label-space restriction, not better prediction: under the five-class restriction the 64 samples the mRNA model sends to the absent claudin-low column are reallocated to in-space classes, which is the same evaluation-rule change quantified in Section~\ref{sec:labelspace}.

We report this as a negative result about modality fusion, not as a failure of the pipeline. The consistent finding is that additional modalities and additional architectural capacity raise internal cross-validated scores while leaving or worsening external accuracy. The one intervention that improved external performance was correcting the label space, not adding features.

\subsection{Label-Space Reconciliation Changes Cross-Cohort Evaluation}
\label{sec:labelspace}

Uncorrected accuracy on TCGA-BRCA (73.76\%) was similar to that on the METABRIC test set (73.8\%), whereas accuracy on GSE21653 was lower (63.16\%). These observations do not support a general claim that label-space mismatch explains every cross-cohort accuracy difference. However, both external cohorts use PAM50 annotations, whose five intrinsic subtypes exclude claudin-low, whereas the METABRIC classifier was trained on six classes. We therefore measured how explicitly reconciling the label space changes the reported metrics, separately from prior correction, without attributing all residual differences to annotation mismatch.

Predictions assigned to a class excluded by the target's annotation standard are necessarily counted as errors under that standard. In TCGA, the model assigned 75 samples (7.6\%) to claudin-low. These predictions included 22.1\% of basal-like and 44.4\% of Normal-like samples. The resulting recall changes depend on which classes the model confuses with claudin-low.

Table~\ref{tab:labelspace} decomposes the correction on both external cohorts. Restricting the output space to the five PAM50 classes and renormalising (mask\_only) already moves TCGA-BRCA from 73.76\% to 79.74\% and GSE21653 from 63.16\% to 68.42\%. Adding EM prior correction over the reconciled space (em\_pam50) moves TCGA-BRCA to \textbf{80.65\%} and GSE21653 to \textbf{74.06\%}, with macro F1 rising from 0.585 to 0.737 and from 0.504 to 0.683 relative to the uncorrected base (though EM itself, applied on top of mask\_only, lowers macro F1 -- see below). On GSE21653 this is a 10.90-point increase relative to the uncorrected estimate. Because the procedure changes both the allowed label set and the estimated class priors, this increase should be interpreted as an evaluation and post-processing effect, not a gain in learned discrimination.

One detail in the table deserves emphasis rather than omission: EM prior correction trades macro F1 for accuracy. On TCGA-BRCA, mask\_only already reaches macro F1 0.757, and the prior correction lowers it to 0.737 while raising accuracy from 79.74\% to 80.65\%. The correction shifts mass toward the dominant class, which can improve overall accuracy at the expense of rarer classes. This is consistent with prior matching and motivates reporting both metrics. The choice between the two rows is therefore a decision about which error matters clinically, not a strictly dominated one; the label-space restriction is unambiguously beneficial (macro F1 0.585 $\rightarrow$ 0.757), while the additional prior correction is a trade-off that a deployment could reasonably decline. The multiclass Brier score, however, improves monotonically along base $\rightarrow$ mask\_only $\rightarrow$ em\_pam50 on both cohorts (TCGA: 0.376 $\rightarrow$ 0.291 $\rightarrow$ 0.275; GSE21653: 0.492 $\rightarrow$ 0.427 $\rightarrow$ 0.378), so the correction produces better probability estimates in a proper-scoring sense rather than merely relabelling the argmax.

\begin{table}[htbp]
\centering
\caption{Decomposition of the cross-cohort correction. Within each cohort, all rows share one classifier and feature set, using logistic regression ($C$=0.01) and within-platform rank normalization. TCGA-BRCA uses 412 genes; GSE21653 uses the 369 genes mapped to GPL570, with a source model refitted on those genes. Rows differ only in label-space handling. \textbf{mask\_only} uses the annotation-standard fact that claudin-low is not a PAM50 subtype; \textbf{em\_pam50} adds EM prior correction on the reconciled space; \textbf{em\_full} runs EM without label-space knowledge. Neither the mask nor any row uses target labels.}
\label{tab:labelspace}
\fitcol{%
\begin{tabular}{lcccc}
\toprule
 & \multicolumn{2}{c}{\textbf{TCGA-BRCA (n=987)}} & \multicolumn{2}{c}{\textbf{GSE21653 (n=266)}} \\
\cmidrule(lr){2-3}\cmidrule(lr){4-5}
\textbf{Estimator} & Acc (\%) & Macro F1 & Acc (\%) & Macro F1 \\
\midrule
base (no reconciliation) & 73.76 & 0.585 & 63.16 & 0.504 \\
mask\_only & 79.74 & 0.757 & 68.42 & 0.626 \\
em\_pam50 & \textbf{80.65} & \textbf{0.737} & \textbf{74.06} & \textbf{0.683} \\
em\_full (no mask) & 75.99 & 0.586 & 68.42 & 0.541 \\
\bottomrule
\end{tabular}}
\end{table}

\begin{figure}[htbp]
\centering
\includegraphics[width=0.62\linewidth]{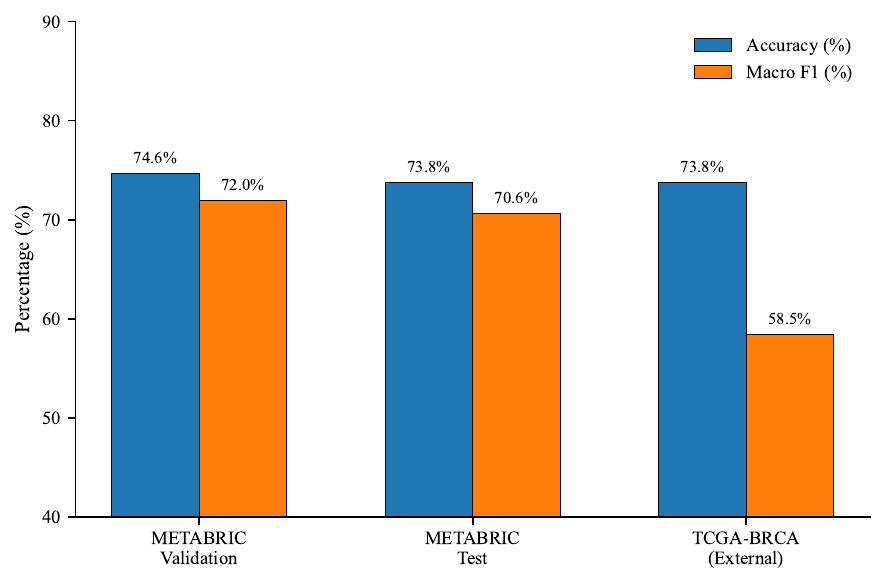}
\caption{Accuracy and macro F1 for the same mRNA-only estimator across METABRIC validation, METABRIC test, and TCGA-BRCA. TCGA accuracy is similar to the internal estimates, while the six-class macro F1 is affected by the absent claudin-low class.}
\label{fig:cross_platform}
\end{figure}

Figure~\ref{fig:cross_platform} compares one mRNA-only estimator across three cohorts. The estimator uses 412 genes, $C$=0.01, and within-platform rank normalization, with no retraining. Accuracy was 74.6\% on METABRIC validation, 73.8\% on the METABRIC test set, and 73.8\% on TCGA-BRCA. Six-class macro F1 was 0.720, 0.706, and 0.585, respectively. For TCGA, averaging over the five PAM50 classes instead gives 0.702. Thus, the macro-F1 difference depends in part on the class set used for scoring; these metrics alone do not establish unchanged model discrimination.

The label-space protocol is a model-agnostic tool, not a property of the logistic regression used above. We re-ran the same harmonised mRNA pipeline (METABRIC train+val, $n$=1{,}163; TCGA-BRCA $n$=694) with three estimator families -- multinomial logistic regression ($C$=0.01), Random Forest (300 trees), and XGBoost ($n_{\text{trees}}$=300) -- and applied the mask to each (Table~\ref{tab:crossmodel}). In every family the mask\_only operation raises external accuracy and removes all predictions into the class the target standard does not annotate. The magnitude of the gain differs by estimator (logistic regression +7.2 points, XGBoost +5.9, Random Forest +3.5), but the direction and the mechanism -- a label-space correction, not a retraining gain -- are consistent. This is the property that makes the protocol integrable as a fixed post-processing module in any classifier's inference stack, independent of the model's inductive bias.

\begin{table}[htbp]
\centering
\caption{Mask-only label-space correction is effective across estimator families on TCGA-BRCA (harmonised mRNA, $n$=694). Base: argmax over the full six-class space. Mask: the absent claudin-low class is dropped from the candidate set and the remaining five re-argmaxed. All estimators are trained on METABRIC train+val ($n$=1{,}163).}
\label{tab:crossmodel}
\fitcol{%
\begin{tabular}{lcccc}
\toprule
\textbf{Estimator} & \textbf{CV acc} & \textbf{Base acc} & \textbf{Mask acc} & $\Delta$ \\
\midrule
Logistic Regression ($C$=0.01) & 72.9\% & 73.34\% & \textbf{80.55}\% & +7.20 \\
Random Forest (300 trees) & 70.4\% & 75.36\% & \textbf{78.82}\% & +3.46 \\
XGBoost (300 trees) & 72.2\% & 72.05\% & \textbf{77.95}\% & +5.91 \\
\bottomrule
\end{tabular}}
\end{table}

The correction changes real patient-level assignments. Of the 694 TCGA-BRCA patients, the uncorrected model predicted the absent claudin-low class for 64 (9.2\%); the 59-dimension clinical block that drives some multimodal variants is irrelevant here, since this is the mRNA-only protocol. Under the five-class PAM50 standard these 64 assignments are errors, and the mask redirects all of them to in-space classes: 34 to Basal-like, 19 to Luminal A, 7 to Normal-like, 3 to HER2-enriched, and 1 to Luminal B. The clinical content of these 64 patients is dominated by receptor-negative disease -- 60 are ER-negative (of whom 60 are ER$^-$/HER2$^-$, i.e. triple-negative-like) and 4 are HER2-positive -- and 31 of the 64 carry a true basal-like label. An un-reconciled pipeline therefore labels 60 ER-negative, largely basal-like patients with a subtype that the target cohort's PAM50 standard does not recognise; the label-space protocol assigns them back to an allowed class, most often basal-like (34 of 64), which is the treatment-relevant (chemotherapy-focused) category for triple-negative disease. At the cohort level this is the entire source of the accuracy gain (73.34\% $\to$ 80.55\%, \textbf{+7.2 points}); at the patient level it is 64 management-relevant reassignments.

\subsection{Unlabelled Predictions Did Not Recover the Label Space}
\label{sec:detector_v2}

We tested whether the absent class could be detected without annotation-space information. Unconstrained EM and two detector procedures did not reliably identify it in the tested settings. \emph{The detection-F1 figures quoted in this and the following two subsections come from the original scoring function, which returns zero when a detector correctly predicts the empty set; every cohort with no absent class was therefore scored as a total failure irrespective of the detector's answer. Section~\ref{sec:detector_v2} re-scores the same simulation under the standard empty-set convention and should be read as the reference for detector ranking.}

Running the same EM procedure without the mask (em\_full), using the same classifier and unlabelled target predictions, leaves claudin-low with an estimated prior of \textbf{0.0751} on TCGA-BRCA and 0.0719 on GSE21653, against a true prevalence of exactly 0. EM has no signal with which to push that mass to zero: a class that is absent from the label space still receives probabilistic predictions, and those predictions are indistinguishable from genuine low-confidence assignments to a rare class. The result is that em\_full recovers only 2.2 of the 6.9 available points on TCGA-BRCA and leaves macro F1 essentially unchanged (0.585 $\rightarrow$ 0.586), whereas the masked version, given one bit of annotation knowledge, recovers 6.9 points. We stress that the two runs differ in nothing else.

In this experiment, the predicted marginal was compatible with both an absent class and a rare class. These results do not prove that no function of unlabeled predictions can distinguish those cases under any model. They show that the annotation-space fact was not recovered by the tested EM procedure and detectors.

We did not accept that conclusion without testing the obvious workaround: detect the absent class geometrically. On the probability simplex, a truly absent class should show a deficit of observed mass relative to its source-expected mass, and we built the ratio statistic described in Section~\ref{sec:methods_detector} to exploit exactly that. We calibrated the threshold on null data, consisting of cohorts with no absent class, at level $\alpha = 0.05$, so that a clean cohort is flagged at most 5\% of the time. This calibration is what makes the test fair; an uncalibrated threshold would flag absence everywhere.

The detector fails. The null-calibrated threshold is $\tau^* = 1.0349$, against a null maximum ratio with median 1.0184 and maximum 1.0493. Because the null distribution of the maximum ratio almost reaches the calibrated threshold, the statistic has almost no power: on the calibration set precision is 0.257 and recall 0.643 (F1 0.367). In the controlled simulation its mean detection F1 is \textbf{0.047} against \textbf{0.549} for EM. Its performance depended on the baseline. Against the uncorrected model it helped on average (mean $+3.34$ points over 30 cells) but was not reliable. Its worst cell is $-2.77$ points, on a cohort with no absent class (73.98\% $\rightarrow$ 71.22\%), precisely because $\tau^*$ flags absent classes that are not there. Against EM, the procedure it was built to replace, it does not help at all: it wins outright in 11 cells, loses in 9, ties in 10, and its mean gain is \textbf{$-0.37$ points}.

We also tested a generative model-selection detector described in Section~\ref{sec:methods_lrt}. It fits the target sample as a Gaussian mixture with source-derived components and selects the label space that minimises BIC over subsets with zero, one, or two absent classes. The complexity penalty replaces a hand-set threshold. On 30 null cohorts with every class present, it selects a non-empty absent set in \textbf{27 of 30 runs (90\%)}, and in 23 of those it claims \emph{two} absent classes. Its mean detection F1 in simulation is 0.209 against 0.549 for EM, and it loses to EM on accuracy in \textbf{all 30 of 30} cells, by a mean of 17.6 points. A BIC-style penalty is not sufficient to make an absent class identifiable.

Both detector procedures produced negative results in the tested settings. The geometric statistic had low detection performance, while the model-selection procedure produced false positives on clean cohorts. These findings support specifying the target label space from cohort documentation rather than relying on the tested detectors.

The comparison above used a scoring function that returns 0.0 whenever a detector correctly predicts the empty set, because the F1 numerator and denominator both vanish; every cohort with no absent class was therefore scored as a total failure regardless of whether the detector was right. Under the standard empty-set convention the comparison changes substantially, so we re-ran the simulation scoring each cell with micro precision, recall, and exact-set match, over 20 seeds per cell, six absent sets, three covariate-shift levels, and with and without a Dirichlet prior shift, giving 720 evaluated cohorts. Raw detected and true sets were written to disk so the scoring can be re-examined without re-running the simulation.

Table~\ref{tab:detector_v2} gives the pooled result. The uncalibrated EM rule that the earlier analysis reported as failing is in fact the strongest of the six on F1 and exact-set match: F1 0.786 and exact-set-match 0.550, both above every calibrated alternative. The multiplicity-controlled replacement, which resamples source validation predictions to build a null and applies Benjamini--Hochberg at $\alpha = 0.05$, is markedly more aggressive: precision falls to 0.577 and it flags a non-empty absent set on 73 of the 120 clean cohorts. The probability-frequency rule flags 87 of 120. Importance weighting the null by a discriminator-estimated density ratio reduces clean-cohort errors to 56 of 120 but also halves recall, buying specificity at a real cost in power. Declaring a class absent because the model never predicts it attains perfect precision and an exact-match rate of 0.167 by predicting nothing, which is the degenerate solution the empty-set convention exposes rather than rewards.

\begin{table}[htbp]
\centering
\caption{Absent-class detectors scored with the standard empty-set convention (20 seeds $\times$ 6 absent sets $\times$ 3 covariate-shift levels $\times$ 2 prior-shift settings $=$ 720 cohorts). Null false positives count cohorts with no absent class on which the detector returned a non-empty set. The na\"ive exact-match rate of 0.167 is the fraction of settings in which nothing was in fact absent.}
\label{tab:detector_v2}
\fitcol{%
\begin{tabular}{lccccc}
\toprule
\textbf{Detector} & \textbf{Micro P} & \textbf{Micro R} & \textbf{F1} & \textbf{Exact} & \textbf{Null FP} \\
\midrule
Uncalibrated EM (\emph{em\_orig}) & \textbf{0.920} & 0.685 & \textbf{0.786} & \textbf{0.550} & 17/120 \\
Calibrated EM (\emph{em\_cal}) & 0.577 & \textbf{0.876} & 0.696 & 0.362 & 73/120 \\
Importance-weighted EM (\emph{em\_cal\_iw}) & 0.540 & 0.544 & 0.542 & 0.242 & 56/120 \\
Probability frequency (\emph{pfreq}) & 0.548 & 0.902 & 0.682 & 0.372 & 87/120 \\
Geometric coherence (\emph{geo}) & 0.467 & 0.154 & 0.232 & 0.193 & 27/120 \\
Never predicted (\emph{naive}) & 1.000 & 0.006 & 0.012 & 0.167 & \textbf{0/120} \\
\bottomrule
\end{tabular}}
\end{table}

The pooled figures conceal the structure that matters: detection degrades with covariate shift and fails differently under prior shift. Restricting to cohorts with a truly absent class, the uncalibrated rule's mean F1 is 0.995 with no covariate shift and no prior shift, falling to 0.543 at the highest covariate-shift level with prior shift present. The calibrated detectors lose control of the clean-cohort error rate when prior shift is present. With no prior shift the uncalibrated rule flags none of the 20 clean cohorts at any covariate-shift level (0/20), while \emph{em\_cal} flags 3, 2, and 11 of 20 and \emph{pfreq} flags 11, 9, and 7 of 20 as the covariate shift rises from 0.0 to 0.8. Under prior shift the uncalibrated rule still flags only 9, 3, and 5 of 20, but \emph{pfreq} flags all 20 clean cohorts at every covariate-shift level and \emph{em\_cal} flags 20, 18, and 19 of 20. No detector is simultaneously high-recall and low-false-positive across both shift axes.

The practical conclusion is unchanged but now rests on corrected measurement rather than a metric artefact. Absence of a class is not identifiable from unlabelled predictions by any tested procedure once covariate and prior shift are both present, and the apparent failure of the original rule was overstated: it is the strongest detector tested on F1 and exact-set match, and its calibrated replacements trade that strength for clean-cohort false positives.

\subsection{Separating Label-Space Mismatch from Prior Shift}
\label{sec:absence_generality}

Because the real cohorts change both prevalence and label space simultaneously, we simulated the two factors independently: label-space mismatch $n_{\text{absent}} \in \{0,1,2\}$, covariate shift, and prior shift, 20 seeds per cell, $n_{\text{target}} = 1000$ (Section~\ref{sec:methods_sim}). The separation is clean. With \emph{no} prior shift the correction is worth $-0.4$ to $+3.2$ points, straddling zero and growing only with the number of absent classes: with no prior shift the label-space term alone contributes roughly $+0.7$ to $+3.2$ points. With prior shift the correction is worth $+5.9$ to $+7.9$ points in every cell. The mechanism is that label-space mismatch creates \emph{apparent} prior shift on the remaining classes. EM can only redistribute mass within the supplied label space, so the two effects can interact in real cohorts.

The simulation also reproduces the identifiability failure at scale: the oracle-mask column matches the detected column only when the mask is supplied, and the unguided detector's accuracy tracks the uncorrected baseline once the absent class is rare.

The real-cohort analysis turns on a single absent class, claudin-low, which is also the rarest in the source prior (9.5\% of METABRIC). A reader may reasonably ask whether the effect is representative of label-space mismatch or merely its smallest instance, so we swept the absence pattern over which class is missing and how it is missing, holding everything else fixed.

Severity is governed by the absent class's \emph{prior mass}, not by its identity. Dropping \textbf{Luminal A} -- the commonest class, at 35.7\% of the source -- costs 11.33 points of naive accuracy, all recoverable ($74.67\% \rightarrow 86.00\%$); dropping claudin-low costs 1.74 points. The class we study is thus the mildest instance, by roughly a factor of six.

More consequential is \emph{how} the class goes missing. Annotation disagreement rarely takes the form of a class disappearing; far more often one class is recorded under a neighbour's label. A Luminal B relabelled as Luminal A -- two subtypes sharing receptor status and differing mainly in proliferation -- leaves naive accuracy at a reassuring 79.43\% while concealing a recoverable loss of \textbf{12.0 points}, the largest in the sweep, and the label-free detector recovers it only partially (detection F1 0.767). Partial absence defeats detection outright for the classes that matter most: when prevalence is scaled to $\tau_{\text{ratio}}$ of source rather than zeroed, detection F1 is \textbf{0.000} for Luminal A and \textbf{0.067} for Luminal B, against 0.967 for claudin-low. The identifiability failure is therefore worst exactly where the consequences are largest, which strengthens the practical recommendation: the target's allowed labels should be read off the cohort documentation, not inferred from predictions.

The correction does not depend on how the target is partitioned. Splitting TCGA-BRCA in half and applying the procedure transductively within each half gives 80.97\% and 80.12\%, a spread of 0.85 points compared with the recovered gain of 6.89. A paired bootstrap over target samples ($n_{\text{iter}} = 2000$) gives a mean gain of \textbf{+6.88 points}, 95\% CI \textbf{[+4.96, +8.81]}, with $P(\text{gain} > 0) = 1.0$. The gain is therefore not an artifact of a favourable partition or of a few influential samples.

\subsection{A Five-Class Training Baseline Reaches the Same Accuracy on TCGA}
\label{sec:fiveclass}

The estimator above reconciles the label space at evaluation time. A simpler
alternative is to train on five classes directly. We built both variants -- dropping
the 91 claudin-low training samples (n=867), and relabelling them as Basal-like
(n=958) -- on the same features and hyperparameters, and compared them with the
six-class estimator on the identical external cohorts (Table~\ref{tab:fiveclass}).

The interpretation we draw is that label-space \emph{reconciliation} accounts for most
of the recovered accuracy on TCGA, whereas prior correction contributes materially only
when the target composition departs from the source. This does not change the reported
result -- 73.76\% remains an artefact of evaluating a six-class model against a
five-class standard -- but it does mean that a five-class training run is a legitimate
design when the target label space is known in advance, and we say so rather than
claiming the reconciliation procedure is uniquely necessary.

\begin{table}[htbp]
\centering
\caption{Five-class training baselines against the six-class estimator with label-space reconciliation. All rows use one rank-normalised mRNA feature set and logistic regression ($C{=}0.01$); rows differ only in training label space and post-hoc handling. Paired McNemar comparisons against the 6-class base are reported in the text; where $p \ge 0.05$ the difference is not statistically distinguishable and is described as such rather than as an advantage.}
\label{tab:fiveclass}
\fitcol{%
\begin{tabular}{llrrrr}
\toprule
\textbf{Cohort} & \textbf{Estimator} & \textbf{Acc (\%)} & \textbf{Macro F1} & \textbf{ECE} & \textbf{Brier} \\
\midrule
    \multicolumn{6}{l}{\textit{TCGA-BRCA} (n=987)} \\
    \quad 6-class base            &       & 73.76 & 0.585 & 0.0352 & 0.3756 \\
    \quad 6-class mask\_only       &       & 79.74 & 0.757 & 0.0557 & 0.2906 \\
    \quad 6-class em\_pam50        &       & \textbf{80.65} & 0.737 & 0.0460 & \textbf{0.2745} \\
    \quad 5-class drop (n=867)     &       & 80.04 & \textbf{0.774} & 0.0560 & 0.2892 \\
    \quad 5-class merge            &       & 76.90 & 0.695 & 0.0344 & 0.3284 \\
    \midrule
    \multicolumn{6}{l}{\textit{GSE21653} (n=266)} \\
    \quad 6-class base            &       & 63.16 & 0.504 & 0.0836 & 0.4920 \\
    \quad 6-class mask\_only       &       & 68.42 & 0.626 & 0.0364 & 0.4271 \\
    \quad 6-class em\_pam50        &       & \textbf{74.06} & 0.683 & 0.0595 & \textbf{0.3776} \\
    \quad 5-class drop (n=867)     &       & 69.55 & \textbf{0.643} & 0.0467 & 0.4025 \\
    \quad 5-class merge            &       & 72.18 & 0.654 & 0.0481 & 0.3747 \\
\bottomrule
\end{tabular}}
\end{table}

On TCGA-BRCA the five-class \emph{drop} model reaches 80.04\% against 80.65\% for the six-class estimator with EM correction (paired McNemar $p=0.50$), and attains the highest macro F1 of any row (0.774). On GSE21653 the six-class estimator with EM correction is significantly better (74.06\% against 69.55\%, $p=0.003$), because EM re-estimates target priors that depart from the source in this basal-enriched cohort; the \emph{merge} variant is worse on TCGA ($p<0.001$) and indistinguishable on GSE21653 ($p=0.23$).

\begin{figure}[htbp]
\centering
\includegraphics[width=0.66\linewidth]{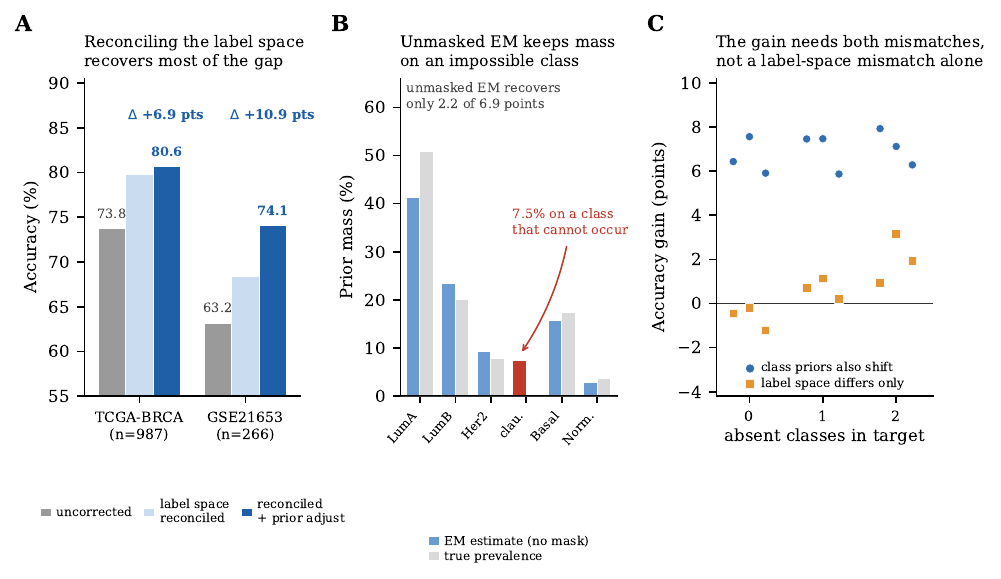}
\caption{Label-space reconciliation and prior correction. \textbf{(A)} Accuracy before and after label-space reconciliation on TCGA-BRCA and GSE21653, showing the mask-only and EM-corrected stages separately. \textbf{(B)} The identifiability failure: unconstrained EM retains substantial prior mass on claudin-low, whose true prevalence in both cohorts is zero, and recovers only a fraction of the available gain. \textbf{(C)} Controlled simulation separating label-space mismatch from prior shift; the correction is near zero without prior shift and consistently large with it.}
\label{fig:label_shift}
\end{figure}

\subsection{Survival Analysis Validates Prognostic Stratification}
\label{sec:survival}
\label{sec:shap}
\label{sec:route_a_treatment}
\label{sec:pcr}
\label{sec:survival}

Cox proportional hazards regression on the METABRIC validation + test sets (n=411, 267 death events, 65\% event rate) showed that predicted subtype probabilities carry statistically significant prognostic information ($p = 0.0008$), though the overall C-Index of 0.598 is modest and comparable to other single-feature prognostic models in breast cancer \cite{prat2014research}. Time-dependent C-Indices were 0.651 (3-year), 0.622 (5-year), and 0.610 (10-year).

\begin{table}[htbp]
\centering
\caption{Cox regression hazard ratios for predicted subtypes (Luminal A = reference) on METABRIC (n=411).}
\label{tab:cox}
\fitcol{%
\begin{tabular}{lccc}
\toprule
\textbf{Predicted Subtype} & \textbf{HR} & \textbf{95\% CI} & \textbf{p-value} \\
\midrule
Luminal B & 1.50 & [1.14, 1.97] & $<0.005$ \\
HER2-enriched & 1.43 & [1.08, 1.89] & 0.01 \\
Basal-like & 1.46 & [1.07, 1.99] & 0.02 \\
Claudin-low & 0.82 & [0.58, 1.16] & 0.25 \\
Normal-like & 1.31 & [0.86, 2.00] & 0.21 \\
\bottomrule
\end{tabular}}
\end{table}

On the TCGA-BRCA cohort (n=966, 133 events), the C-Index was 0.577 ($p = 0.047$), with time-dependent C-Indices of 0.587 (3-year), 0.574 (5-year), and 0.577 (10-year). The concordant survival discrimination across both cohorts (0.598 internally, 0.577 and 0.584 externally) is modest but consistent, suggesting the risk signal generalizes across platforms despite the limited discriminative power. The risk score is a Cox model fit on METABRIC training predictions and applied unchanged to TCGA, which requires no outcome information from the validation cohort and is therefore fully reproducible from the released code. An earlier iteration of this analysis reported a C-index of 0.586 ($p = 0.013$) under a variant risk-score construction whose exact form was not retained; because that construction cannot be reconstructed from the preserved pipeline, we report the reproducible figures throughout rather than mixing the two.

\begin{figure}[htbp]
\centering
\includegraphics[width=0.62\linewidth]{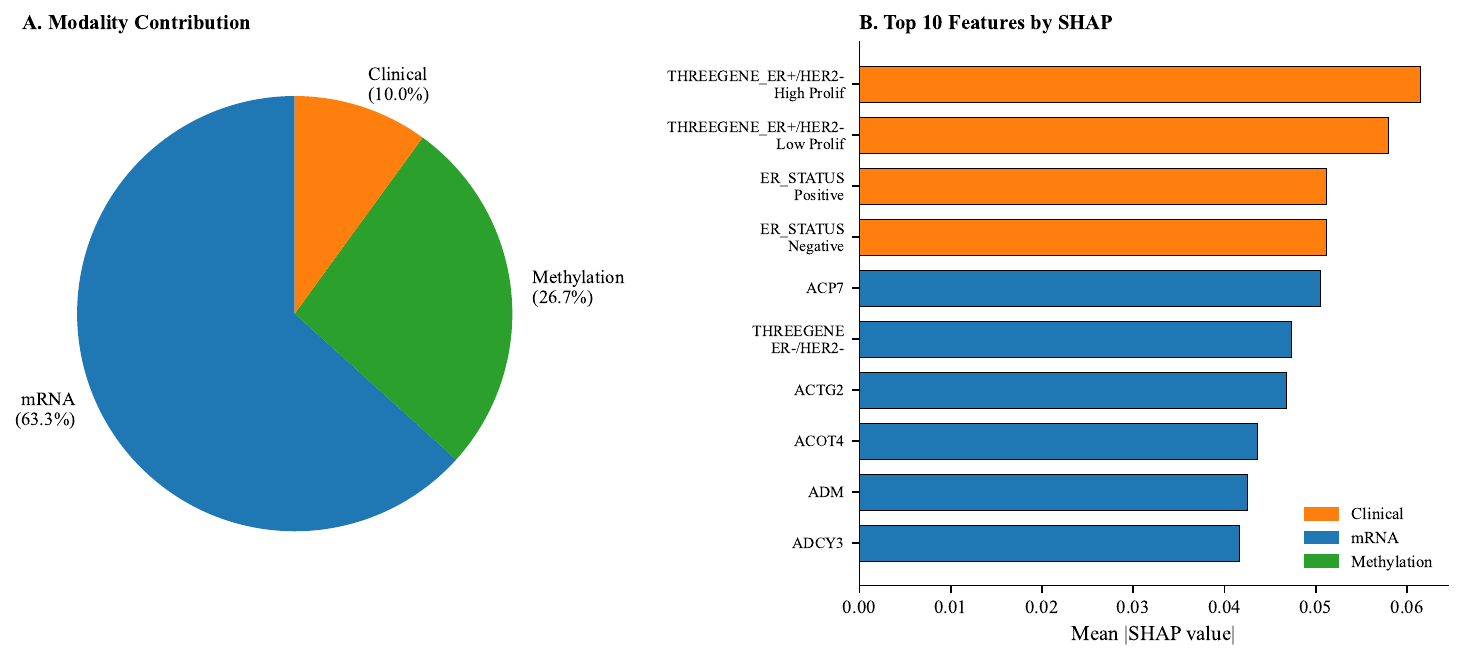}
\caption{SHAP-based modality attribution. mRNA expression accounts for 63.3\% of model decisions, methylation 26.7\% and clinical features 10.0\%; copy-number and mutation blocks contribute under 1\% each.}
\label{fig:shap}
\end{figure}

SHAP interpretability analysis (Figure~\ref{fig:shap}) quantified the relative importance of each modality. Because the classifier is linear, these attributions are exact and sum to the model's output, so they can be read as the model's own decomposition of its decisions. mRNA expression contributed 63.3\%, methylation 26.7\% and clinical features 10.0\%; copy number and mutation data contributed $<$1\% each, consistent with their noise-inducing effect in the ablation. These figures describe \emph{this} fitted model on \emph{this} encoding; they are not evidence that the two weak modalities carry no biological signal, only that the present architecture and feature representation extract little from them.

The top discriminating features included ER/HER2 clinical markers and proliferation features. Leading mRNA features included \textit{ACP7}, \textit{ACTG2}, \textit{ADIRF}, and \textit{ACOT4}. These features align with established breast cancer biology and PAM50 subtype definitions.

We also asked whether the 713-dimension panel adds information beyond the subtype call itself for clinically relevant stratified outcomes. We defined a 5-year outcome (death within 60 months versus survival beyond it, censored observations excluded) and evaluated within treatment subgroups, using the same 5-fold protocol and reporting AUC against a PAM50 subtype-only baseline trained on the identical splits.

A necessary caveat: METABRIC's treatment fields record whether a therapy was \emph{administered}, not whether the patient responded. There is no response variable in this dataset, so this is \emph{prognosis among treated patients}, not treatment-response prediction, and we label it as such throughout.

\begin{table}[htbp]
\centering
\caption{5-year outcome discrimination within treatment subgroups, genomic panel versus PAM50 subtype alone. These are administered-treatment cohorts, not response assessments.}
\label{tab:treatment}
\fitcol{%
\begin{tabular}{lccccc}
\toprule
\textbf{Treated subgroup} & \textbf{n} & \textbf{Events} & \textbf{AUC (panel)} & \textbf{AUC (subtype)} & \textbf{$\Delta$} \\
\midrule
Chemotherapy & 275 & 120 & 0.624 & 0.571 & $+0.053$ \\
Hormone therapy & 768 & 210 & 0.679 & 0.577 & $+0.102$ \\
Radiotherapy & 767 & 225 & 0.718 & 0.640 & $+0.078$ \\
\bottomrule
\end{tabular}}
\end{table}

The panel out-discriminates the subtype-only baseline in all three subgroups (Table~\ref{tab:treatment}), with the largest margin in the hormone-therapy cohort ($+0.102$). Because a subtype-only baseline already encodes the dominant prognostic axis, an increment of this size is consistent with additional prognostic information in within-subtype expression and methylation variation. This comparison favours the panel over the subtype-only baseline for risk stratification in these METABRIC treatment subgroups and does not depend on the Transformer.

Subtype accuracy is a surrogate endpoint. To test whether the framework carries clinical value we applied it to a genuine response endpoint: pathological complete response (pCR) after neoadjuvant chemotherapy in GSE20194 (n=278, 56 pCR / 222 residual disease, Affymetrix HG-U133A). The 412-gene panel covers 276 genes on this platform (67.0\%). We compared a clinical-only model (age, ER, PR, HER2, grade, histology) against the same model augmented with genomic features and with predicted subtype probabilities.

The genomic panel did not significantly improve on clinical features in this cohort. Clinical-only AUC was 0.791, genomic-only AUC was 0.773, and combined AUC was 0.794. The difference was \textbf{+0.0045} (95\% CI \textbf{[$-0.060$, $+0.061$]}, $P(\Delta > 0) = 0.576$). Adding predicted subtype probabilities to the clinical model gave an AUC of 0.797, a difference of $+0.0034$ (95\% CI [$-0.023$, $+0.030$]). Neither increment was distinguishable from zero. The genomic-only model had lower AUC than the clinical-only model.

What the predictions do capture is biology. Grouping patients by predicted subtype gives pCR rates that followed the observed subtype pattern: HER2-enriched 53.6\% (15/28), Basal-like 42.4\% (14/33), claudin-low 26.2\% (11/42), Luminal B 13.2\% (9/68), Luminal A 6.0\% (6/100), Normal-like 14.3\% (1/7). HER2-enriched and Basal-like had higher observed pCR rates than Luminal A. The Luminal A estimate is based on 100 patients. Predicted-subtype probabilities alone reached AUC 0.759, comparable to ER status alone (0.727).

We report this as a negative result. In this cohort, the genomic panel did not improve pCR prediction beyond routine clinical variables, including receptor status and grade. Its observed value in this study was prognostic rather than predictive. This is consistent with the related analysis in Section~\ref{sec:route_a_treatment}, where the panel out-discriminates a subtype-only baseline for \emph{outcome} within treated subgroups while METABRIC records treatment administration rather than response.

\begin{figure}[htbp]
\centering
\includegraphics[width=0.66\linewidth]{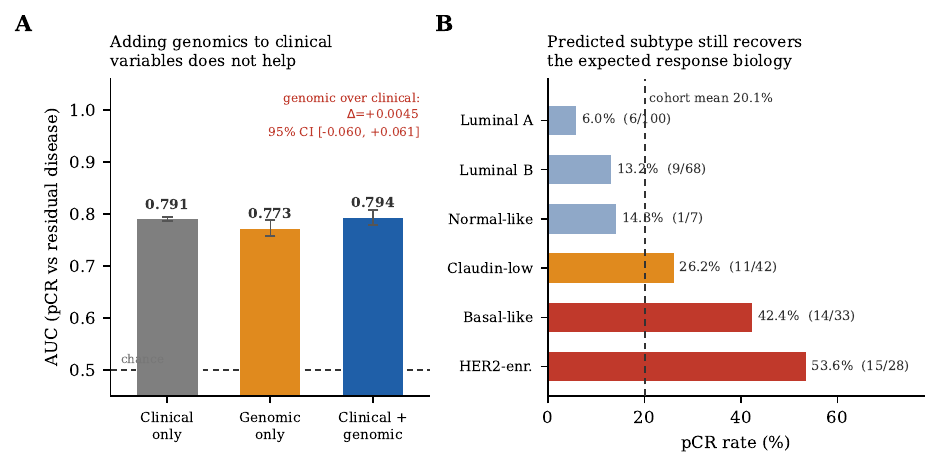}
\caption{Treatment-response prediction in GSE20194 (n=278; 56 pCR, 222 residual disease). \textbf{(A)} Discrimination of pCR after neoadjuvant chemotherapy. The genomic panel does not improve on clinical variables: the increment is $+0.0045$ with a 95\% CI spanning zero, and the genomic-only model (0.773) is worse than clinical-only (0.791). \textbf{(B)} pCR rate by predicted subtype. Despite adding no predictive value over clinical features, the predicted subtype reproduces the observed response pattern, with pCR rates of 53.6\% for HER2-enriched and 42.4\% for basal-like compared with 6.0\% for Luminal A.}
\label{fig:pcr}
\end{figure}

\section{Discussion}
\label{sec:discussion}

Our primary contribution is the label-space compliance protocol: an empirical analysis of how annotation-space differences affect cross-cohort evaluation of breast cancer molecular-subtype classifiers, formalized as an evaluation protocol, with a concrete clinical corollary for decision-support systems built on subtyping. Applying an externally specified PAM50 label space changes reported scores for two cohorts -- TCGA-BRCA accuracy moves from 73.76\% to 80.65\% and GSE21653 from 63.16\% to 74.06\% after restricting the label space and correcting priors -- and this change must be distinguished from improving the classifier itself. Because these operations alter evaluation rules, not representation, a subtyping model's reported accuracy against a cohort whose annotation standard differs from the training space must not be read as model skill.

\textbf{The label space is not recoverable from unlabelled predictions.} Unconstrained EM retains an estimated prior of 0.0751 for the absent claudin-low class on TCGA-BRCA, and neither a calibrated geometric detector (simulated detection F1 0.047--0.232) nor a generative model-selection procedure (selects a non-empty absent set in 90\% of clean runs) recovers the target annotation space under covariate and prior shift. The practical consequence is that the target label set must be specified from cohort documentation, not inferred from predictions; this is a prerequisite for valid cross-cohort comparison rather than an optional refinement.

\textbf{The multimodal advantage is internal, and the portable signal is survival.} The three-modality model that wins on METABRIC depends on 59 dataset-specific clinical dimensions with no external counterpart; in a harmonized three-layer space shared with TCGA-BRCA (n=694), no configuration beats mRNA alone (McNemar p=0.13--1.00), and the cross-modal fusion Transformer did not reach the linear reference even with capacity sweeps, regularization, and pretraining. What transfers cleanly across all three cohorts is the survival risk signal (C-index 0.577--0.584 externally, 0.598 internally). For clinical deployment, modality and architecture choices validated on a single cohort are weak evidence of cross-cohort value; we recommend reporting the model-validity envelope -- the annotation space, feature schema, and cohort mix for which a model was developed -- alongside any accuracy figure, and treating survival discrimination rather than subtype accuracy as the portable endpoint.

\textbf{Relationship to established assays and clinical tools.} We do not claim superiority over Prosigna or other locked PAM50 assays: these are not publicly executable, and our external scores measure agreement with the originating studies' PAM50 calls, whose inter-platform concordance is itself imperfect, particularly at the Luminal A/B boundary. Predicted subtypes carry significant but modest prognostic stratification (C-index 0.598, p=0.0008 in METABRIC; 0.577--0.584 externally), below established tools such as Oncotype DX, and the genomic panel adds nothing over clinical variables for pathological complete response (DeltaAUC = +0.0045, CI spanning zero). The value of the framework lies in standardizing how annotation standards are reconciled across cohorts, not in outperforming clinical benchmarks.

\textbf{Limitations.} The model was developed on METABRIC. We assessed ancestry-stratified transfer of the protocol on TCGA-BRCA: the mask correction gained +7.4 points in White (n=487) and +6.6 points in Black/African-American (n=136) patients -- a comparable gain in both groups -- while the Asian subgroup (n=36) was too small to evaluate reliably (one reclassifiable patient). Subtype composition differs by ancestry, with basal-like enrichment among Black patients, consistent with published population differences; generalisation to other ancestries and prospective cohorts still requires external validation. The real-cohort analysis covers one absent class (claudin-low) in two cohorts; the simulation generalizes to every class and absence mode but uses Gaussian-mixture surrogates and is not a forecast for a named cohort. The multimodal model was evaluated externally only in a reduced shared space, and the training sample size may constrain nonlinear models. These conclusions are bounded by the studied cohorts, label spaces, and model configurations and should not be generalized without further validation.
\section{Conclusion}

We have shown that cross-cohort subtype metrics are not comparable until the evaluation label space is reconciled with the target cohort's annotation standard. On TCGA-BRCA, reconciling the six-class training space to the five PAM50 classes and correcting priors moves accuracy from 73.76\\% to 80.65\\%; on GSE21653, from 63.16\\% to 74.06\\%. These are changes in evaluation rules, not in the learned representation, and unconstrained EM (estimated prior 0.0751 on the absent claudin-low class) and the tested detectors could not recover the target label space from unlabelled predictions. The same logic applies to every clinical decision-support system built on subtyping: a model's reported performance against a cohort whose annotation standard differs from the training space is an artefact of that mismatch and must not be read as model skill.

The clinical corollary is threefold. First, cohort documentation, not prediction statistics, should define the allowed label set before any comparative metric is computed. Second, the portable result across platforms is survival discrimination (C-index 0.577--0.584 externally, 0.598 internally) rather than subtype accuracy, which is entangled with label-space and prevalence effects. Third, internal-only evidence for multimodal fusion or architectural capacity is weak: in a harmonized space (n=694) no configuration beat mRNA alone (McNemar p=0.13--1.00), and the best internal model depended on 59 datasetspecific clinical dimensions. For healthcare information systems, this supports reporting the model-validity envelope--the annotation space, feature schema, and cohort mix for which the model was developed--alongside any accuracy figure.

These conclusions are bounded by the studied cohorts, label spaces, and model configurations and should not be generalized without further validation.
\section*{Declarations}
\textbf{Data availability:} METABRIC and TCGA-BRCA are available through cBioPortal (\url{https://www.cbioportal.org/}). GEO accession records are available for GSE21653 and GSE20194. Redistribution of processed data should follow the terms of the source repositories.

\textbf{Code availability:} The code and preprocessed data used for the analyses in this manuscript are publicly available at \href{https://github.com/HAHA12345677888/pam50-label-space}{https://github.com/HAHA12345677888/pam50-label-space}.

\textbf{Author contributions:} Suxing Liu contributed to the study conception, data analysis, and manuscript preparation. Zhijian Zheng and Jiahao Wang contributed to the analysis and to critical revision of the manuscript. All authors read and approved the final manuscript.

\textbf{Funding:} This research received no external funding.

\textbf{Competing interests:} The authors declare no competing interests.

\textbf{Ethics approval and consent to participate:} The analyses use public cohort resources. The analyses were conducted in compliance with institutional review requirements for publicly available datasets.

\textbf{Correspondence:} Correspondence and requests for materials should be addressed to Suxing Liu (email: liushuxing8888@student.usm.my).

\bibliographystyle{IEEEtran}
\bibliography{references}

\begin{thebibliography}{10}
\providecommand{\url}[1]{#1}
\csname url@samestyle\endcsname
\providecommand{\newblock}{\relax}
\providecommand{\bibinfo}[2]{#2}
\providecommand{\BIBentrySTDinterwordspacing}{\spaceskip=0pt\relax}
\providecommand{\BIBentryALTinterwordstretchfactor}{4}
\providecommand{\BIBentryALTinterwordspacing}{\spaceskip=\fontdimen2\font plus
\BIBentryALTinterwordstretchfactor\fontdimen3\font minus \fontdimen4\font\relax}
\providecommand{\BIBforeignlanguage}[2]{{%
\expandafter\ifx\csname l@#1\endcsname\relax
\typeout{** WARNING: IEEEtran.bst: No hyphenation pattern has been}%
\typeout{** loaded for the language `#1'. Using the pattern for}%
\typeout{** the default language instead.}%
\else
\language=\csname l@#1\endcsname
\fi
#2}}
\providecommand{\BIBdecl}{\relax}
\BIBdecl

\bibitem{perou2000molecular}
C.~M. Perou, T.~S{\o}rlie, M.~B. Eisen, M.~van~de Rijn, S.~S. Jeffrey, C.~A. Rees, J.~R. Pollack, D.~T. Ross, H.~Johnsen, L.~A. Akslen \emph{et~al.}, ``Molecular portraits of human breast tumours,'' \emph{Nature}, vol. 406, no. 6797, pp. 747--752, 2000.

\bibitem{parker2009supervised}
J.~S. Parker, M.~Mullins, M.~C. Cheang, S.~Leung, D.~Voduc, T.~Vickery, S.~Davies, C.~Fauron, X.~He, Z.~Hu \emph{et~al.}, ``Supervised risk predictor of breast cancer based on intrinsic subtypes,'' \emph{Journal of Clinical Oncology}, vol.~27, no.~8, pp. 1160--1167, 2009.

\bibitem{prat2010phenotypic}
A.~Prat, J.~S. Parker, O.~Karginova, C.~Fan, C.~Livasy, J.~I. Herschkowitz, X.~He, and C.~M. Perou, ``Phenotypic and molecular characterization of the claudin-low intrinsic subtype of breast cancer,'' \emph{Breast Cancer Research}, vol.~12, no.~5, pp. 1--18, 2010.

\bibitem{prat2014research}
A.~Prat, G.~Bianchini, M.~Thomas, A.~Belousov, M.~C. Cheang, A.~Koehler, P.~G{\'o}mez, V.~Semiglazov, W.~Eiermann, S.~Tjulandin \emph{et~al.}, ``Research-based {PAM50} subtype predictor identifies higher responses and improved survival outcomes in {HER2}-positive breast cancer in the {NOAH} study,'' \emph{Clinical Cancer Research}, vol.~20, no.~2, pp. 511--521, 2014.

\bibitem{haybittle1982prognostic}
J.~Haybittle, R.~Blamey, C.~Elston, J.~Johnson, P.~Doyle, F.~Campbell, R.~Nicholson, and K.~Griffiths, ``A prognostic index in primary breast cancer,'' \emph{British Journal of Cancer}, vol.~45, no.~3, pp. 361--366, 1982.

\bibitem{pasculli2018epigenetics}
B.~Pasculli, R.~Barbano, and P.~Parrella, ``Epigenetics of breast cancer: Biology and clinical implication in the era of precision medicine,'' \emph{Seminars in Cancer Biology}, vol.~51, pp. 22--35, 2018.

\bibitem{goldhirsch2013personalizing}
A.~Goldhirsch, E.~Winer, A.~Coates, R.~Gelber, M.~Piccart-Gebhart, B.~Th{\"u}rlimann, H.-J. Senn, and P.~members, ``Personalizing the treatment of women with early breast cancer: highlights of the st gallen international expert consensus on the primary therapy of early breast cancer 2013,'' \emph{Annals of Oncology}, vol.~24, no.~9, pp. 2206--2223, 2013.

\bibitem{wang2014similarity}
B.~Wang, A.~M. Mezlini, F.~Demir, M.~Fiume, Z.~Tu, M.~Brudno, B.~Haibe-Kains, and A.~Goldenberg, ``Similarity network fusion for aggregating data types on a genomic scale,'' \emph{Nature Methods}, vol.~11, no.~3, pp. 333--337, 2014.

\bibitem{saerens2002adjusting}
M.~Saerens, P.~Latinne, and C.~Decaestecker, ``Adjusting the outputs of a classifier to new a priori probabilities: a simple procedure,'' \emph{Neural Computation}, vol.~14, no.~1, pp. 21--41, 2002.

\bibitem{moon2023machine}
I.~Moon, J.~LoPiccolo, S.~C. Baca, L.~M. Sholl, K.~L. Kehl, M.~J. Hassett, D.~Liu, D.~Schrag, and A.~Gusev, ``Machine learning for genetics-based classification and treatment response prediction in cancer of unknown primary,'' \emph{Nature Medicine}, vol.~29, no.~8, pp. 2057--2067, 2023.

\bibitem{jiang2024autosurv}
L.~Jiang, C.~Xu, Y.~Bai, A.~Liu, Y.~Gong, Y.~Wang, and H.~Deng, ``Autosurv: interpretable deep learning framework for cancer survival analysis incorporating clinical and multi-omics data,'' \emph{npj Precision Oncology}, vol.~8, no.~1, 2024.

\bibitem{lin2026moagnn}
C.~Lin, Y.~Ho, Y.~Chiu, Y.~Tang, Y.~S. Paik, G.~Chen, W.~Huang, and T.~Lee, ``Moagnn: a multi-omics hierarchical graph neural network for subtype classification and prognosis prediction in lung adenocarcinoma,'' \emph{Briefings in Bioinformatics}, vol.~27, no.~1, 2026.

\bibitem{fan2026cmgl}
B.~Fan, H.~Yin, S.~Yi, Y.~Wang, Z.~Li, L.~Zhou, J.~Lv, and W.~Ju, ``Cmgl: Confidence-guided multi-omics graph learning for cancer subtype classification,'' \emph{Bioinformatics}, 2026.

\bibitem{tan2025amogel}
C.~Y. Tan, H.~F. Ong, C.~H. Lim, M.~S. Tan, E.~H. Ooi, and K.~Wong, ``Amogel: a multi-omics classification framework using associative graph neural networks with prior knowledge for biomarker identification,'' \emph{BMC Bioinformatics}, vol.~26, no.~1, 2025.

\bibitem{dou2025mogcan}
Y.~Dou and G.~Mirzaei, ``Mo-gcan: multi-omics integration based on graph convolutional and attention networks,'' \emph{Bioinformatics}, vol.~41, no.~8, 2025.

\bibitem{zhang2025hallmarkgraph}
Q.~Zhang, F.~Liu, and X.~Lai, ``Hallmarkgraph: a cancer hallmark informed graph neural network for classifying hierarchical tumor subtypes,'' \emph{Bioinformatics}, vol.~41, no.~9, 2025.

\bibitem{tanvir2024mogat}
R.~B. Tanvir, M.~M. Islam, M.~Sobhan, D.~Luo, and A.~M. Mondal, ``Mogat: A multi-omics integration framework using graph attention networks for cancer subtype prediction,'' \emph{International Journal of Molecular Sciences}, vol.~25, no.~5, p. 2788, 2024.

\bibitem{xiao2023graph}
S.~Xiao, H.~Lin, C.~Wang, S.~Wang, and J.~C. Rajapakse, ``Graph neural networks with multiple prior knowledge for multi-omics data analysis,'' \emph{IEEE Journal of Biomedical and Health Informatics}, vol.~27, no.~9, pp. 4591--4600, 2023.

\bibitem{zhang2025multiomics}
Y.~Zhang, C.~Yan, Z.~Yang, M.~Zhou, and J.~Sun, ``Multi-omics deep-learning prediction of homologous recombination deficiency-like phenotype improved risk stratification and guided therapeutic decisions in gynecological cancers,'' \emph{IEEE Journal of Biomedical and Health Informatics}, vol.~29, no.~3, pp. 1861--1871, 2025.

\bibitem{yang2024multimodal}
H.~Yang, M.~Yang, J.~Chen, G.~Yao, Q.~Zou, and L.~Jia, ``Multimodal deep learning approaches for precision oncology: a comprehensive review,'' \emph{Briefings in Bioinformatics}, vol.~26, no.~1, 2024.

\bibitem{zhang2024msfn}
G.~Zhang, C.~Ma, C.~Yan, H.~Luo, J.~Wang, W.~Liang, and J.~Luo, ``Msfn: a multi-omics stacked fusion network for breast cancer survival prediction,'' \emph{Frontiers in Genetics}, vol.~15, 2024.

\bibitem{liu2023attentionbased}
J.~Liu, D.~Capurro, A.~Nguyen, and K.~Verspoor, ``Attention-based multimodal fusion with contrast for robust clinical prediction in the face of missing modalities,'' \emph{Journal of Biomedical Informatics}, vol. 145, p. 104466, 2023.

\bibitem{benkirane2023customics}
H.~Benkirane, Y.~Pradat, S.~Michiels, and P.~Courn{\`{e}}de, ``Customics: A versatile deep-learning based strategy for multi-omics integration,'' \emph{PLOS Computational Biology}, vol.~19, no.~3, p. e1010921, 2023.

\bibitem{zhang2025attentionbased}
W.~Zhang, S.~Zhang, J.~You, F.~Li, X.~Wu, X.~Lu, Q.~Lv, J.~Huang, Y.~Yi, and H.~Bu, ``Attention-based multimodal fusion transformer for predicting the efficacy of neoadjuvant therapy in breast cancer: a cross-institutional retrospective study,'' \emph{Breast Cancer Research}, vol.~28, no.~1, 2025.

\bibitem{mao2025a}
N.~Mao, Y.~Dai, H.~Zhou, F.~Lin, T.~Zheng, Z.~Li, P.~Yang, F.~Zhao, Q.~Li, W.~Wang, Y.~Liang, H.~Xie, H.~Ma, L.~Zhang, Y.~Guo, X.~Song, H.~Zhang, and J.~Lu, ``A multimodal and fully automated system for prediction of pathological complete response to neoadjuvant chemotherapy in breast cancer,'' \emph{Science Advances}, vol.~11, no.~18, 2025.

\bibitem{cheng2026aligned}
Y.~Cheng, Y.~Su, Y.~Fan, Y.~Yang, X.~Chen, F.~Wang, K.~Wong, and X.~Li, ``Aligned cross-modal integration and regulatory heterogeneity characterization of single-cell multiomic data with deep contrastive learning,'' \emph{Genome Medicine}, vol.~18, no.~1, 2026.

\bibitem{jia2026dynamic}
S.~Jia, A.~Lysenko, K.~A. Boroevich, A.~Sharma, and T.~Tsunoda, ``Dynamic multimodal survival prediction in multiple myeloma integrating gene expression, longitudinal laboratory measurements, and treatment history,'' \emph{Briefings in Bioinformatics}, vol.~27, no.~5, 2026.

\bibitem{steyaert2023multimodal}
S.~Steyaert, M.~Pizurica, D.~Nagaraj, P.~Khandelwal, T.~Hernandez-Boussard, A.~J. Gentles, and O.~Gevaert, ``Multimodal data fusion for cancer biomarker discovery with deep learning,'' \emph{Nature Machine Intelligence}, vol.~5, no.~4, pp. 351--362, 2023.

\bibitem{kibria2026decoding}
M.~R. Kibria, S.~Lafond, and J.~Arslan, ``Decoding the multimodal maze: A systematic review on the adoption of explainability in multimodal attention-based models,'' \emph{Information Fusion}, vol. 135, p. 104405, 2026.

\bibitem{stan2024unsupervised}
S.~Stan and M.~Rostami, ``Unsupervised model adaptation for source-free segmentation of medical images,'' \emph{Medical Image Analysis}, vol.~95, p. 103179, 2024.

\bibitem{yang2025a1}
H.~Yang, X.~Chen, and C.~Ren, ``A generalized label shift perspective for cross-domain gaze estimation,'' in \emph{Advances in Neural Information Processing Systems 38}, 2025, pp. 5780--5805.

\bibitem{yoshida2025classprior}
T.~Yoshida, E.~Shin'ya, and T.~Washio, ``Class-prior probability estimation using density ratio between unlabeled instances and positively labeled noisy instances,'' \emph{Neurocomputing}, vol. 618, p. 129074, 2025.

\bibitem{volpi2025leap}
L.~Volpi, A.~Moreo, and F.~Sebastiani, ``Leap: Linear equations for classifier accuracy prediction under prior probability shift,'' \emph{Machine Learning}, vol. 114, no.~12, 2025.

\bibitem{chen2025vpsfda}
Y.~Chen, Y.~Wang, and Z.~Xie, ``Vp-sfda: Visual prompt source-free domain adaptation for cross-modal medical image,'' \emph{Health Data Science}, vol.~5, 2025.

\bibitem{lai2023domainaware}
H.~Lai, Y.~Luo, B.~Li, G.~Zhang, and J.~Lu, ``Domain-aware dual attention for generalized medical image segmentation on unseen domains,'' \emph{IEEE Journal of Biomedical and Health Informatics}, vol.~27, no.~5, pp. 2399--2410, 2023.

\bibitem{sayed2026immunohistochemistry}
S.~Sayed, L.~Feng, H.~Koka, M.~H. Mansour, D.~Wang, M.~Brown, K.~Jones, M.~Maina, V.~Ngundo, A.~M. Githuka, Z.~Ajuoga, S.~Fan, A.~N. Hurson, X.~Hua, P.~Lenz, R.~W. Kamau, B.~A. Ooro, F.~Makokha, S.~Ambs, and A.~M. Rozeboom, ``Immunohistochemistry vs molecular breast cancer subtyping among african populations,'' \emph{JAMA Network Open}, vol.~9, no.~10, p. e2638664, 2026.

\bibitem{jrgensen2021pam}
C.~L.~T. J{\o}rgensen, A.~Larsson, C.~Forsare, K.~Aaltonen, S.~Jansson, R.~Bradshaw, P.~Bendahl, and L.~Ryd{\'{e}}n, ``Pam50 intrinsic subtype profiles in primary and metastatic breast cancer show a significant shift toward more aggressive subtypes with prognostic implications,'' \emph{Cancers}, vol.~13, no.~7, p. 1592, 2021.

\bibitem{hohmann2025molecular}
L.~Hohmann, D.~F. Nacer, M.~Aine, Y.~Memari, D.~Black, R.~Bowden, H.~R. Davies, {\AA}.~Borg, J.~Vallon-Christersson, S.~Nik-Zainal, and J.~Staaf, ``Molecular profiling of the basal-like intrinsic molecular subtype in primary er-positive her2-negative breast cancer,'' \emph{Genome Medicine}, vol.~17, no.~1, 2025.

\bibitem{hurson2023reproducibility}
A.~N. Hurson, A.~M. Hamilton, L.~T. Olsson, E.~L. Kirk, M.~E. Sherman, B.~C. Calhoun, J.~Geradts, and M.~A. Troester, ``Reproducibility and intratumoral heterogeneity of the pam50 breast cancer assay,'' \emph{Breast Cancer Research and Treatment}, vol. 199, no.~1, pp. 147--154, 2023.

\bibitem{kandan2025rank}
S.~U. Kandan and O.~Abul, ``Rank normalization of pam50 gene expression for robust breast cancer subtype classification,'' in \emph{2025 17th International Conference on Bioinformatics and Biomedical Technology (ICBBT)}, 2025, pp. 130--134.

\bibitem{rajkumar2025enhanced}
P.~Raj-Kumar, B.~Chen, M.~Hu, T.~A. Hohenstein, J.~Liu, C.~D. Shriver, X.~Lin, and H.~Hu, ``Enhanced pam50 subtyping of breast cancer implemented in the pcapam50 r package,'' \emph{Scientific Reports}, vol.~16, no.~1, 2025.

\bibitem{denegreirosbotan2026corepam}
R.~de~Negreiros~Botan and J.~B. de~Sousa, ``Corepam: a 24-gene pam50-derived expression score with cross-platform external validation for breast cancer prognosis,'' \emph{Breast Cancer Research}, vol.~28, no.~1, 2026.

\bibitem{rojasvelazquez2024methodology}
D.~Rojas-Velazquez, S.~Kidwai, A.~D. Kraneveld, A.~Tonda, D.~Oberski, J.~Garssen, and A.~Lopez-Rincon, ``Methodology for biomarker discovery with reproducibility in microbiome data using machine learning,'' \emph{BMC Bioinformatics}, vol.~25, no.~1, 2024.

\bibitem{van2024a}
R.~Van, D.~Alvarez, T.~Mize, S.~Gannavarapu, L.~Chintham~Reddy, F.~Nasoz, and M.~V. Han, ``A comparison of rna-seq data preprocessing pipelines for transcriptomic predictions across independent studies,'' \emph{BMC Bioinformatics}, vol.~25, no.~1, 2024.

\bibitem{ng2023the}
S.~Ng, S.~Masarone, D.~Watson, and M.~R. Barnes, ``The benefits and pitfalls of machine learning for biomarker discovery,'' \emph{Cell and Tissue Research}, vol. 394, no.~1, pp. 17--31, 2023.

\bibitem{curtis2012genomic}
C.~Curtis, S.~P. Shah, S.-F. Chin, G.~Turashvili, O.~M. Rueda, M.~J. Dunning, D.~Speed, A.~G. Lynch, S.~Samarajiwa, Y.~Yuan \emph{et~al.}, ``The genomic and transcriptomic architecture of 2,000 breast tumours reveals novel subgroups,'' \emph{Nature}, vol. 486, no. 7403, pp. 346--352, 2012.

\bibitem{cancer2012comprehensive}
C.~G.~A. Network, ``Comprehensive molecular portraits of human breast tumours,'' \emph{Nature}, vol. 490, no. 7418, pp. 61--70, 2012.

\bibitem{pedregosa2011scikit}
F.~Pedregosa, G.~Varoquaux, A.~Gramfort, V.~Michel, B.~Thirion, O.~Grisel, M.~Blondel, P.~Prettenhofer, R.~Weiss, V.~Dubourg \emph{et~al.}, ``Scikit-learn: Machine learning in python,'' \emph{Journal of Machine Learning Research}, vol.~12, pp. 2825--2830, 2011.

\bibitem{vaswani2017attention}
A.~Vaswani, N.~Shazeer, N.~Parmar, J.~Uszkoreit, L.~Jones, A.~N. Gomez, {\L}.~Kaiser, and I.~Polosukhin, ``Attention is all you need,'' in \emph{Advances in Neural Information Processing Systems}, vol.~30, 2017, pp. 5998--6008.

\bibitem{lundberg2017unified}
S.~M. Lundberg and S.-I. Lee, ``A unified approach to interpreting model predictions,'' \emph{Advances in Neural Information Processing Systems}, vol.~30, pp. 4765--4774, 2017.

\end{thebibliography}

\end{document}